\documentclass{article}

\usepackage{microtype}
\usepackage{graphicx}
\usepackage{subcaption}
\usepackage{booktabs} % for professional tables
\usepackage{pifont} % for checkmark/cross symbols
\usepackage[table]{xcolor} % for colored table cells
\usepackage{float} % for [H] float placement

\usepackage{hyperref}
\usepackage{xurl}        % better URL line breaking
\hypersetup{breaklinks=true}

\usepackage[accepted]{icml2025}
\makeatletter
\renewcommand{\ICML@appearing}{Preprint.}
\makeatother

\usepackage{amsmath}
\usepackage{amssymb}
\usepackage{mathtools}
\usepackage{amsthm}

\usepackage[capitalize,noabbrev]{cleveref}

\usepackage{listings}
\theoremstyle{plain}

\theoremstyle{definition}

\theoremstyle{remark}

\newcommand{\name}{SocialGrid~}

\icmltitlerunning{SocialGrid: A Benchmark for Planning and Social Reasoning in Embodied Multi-Agent Systems}

\begin{document}

\twocolumn[
    \icmltitle{SocialGrid: A Benchmark for Planning and Social Reasoning \\in Embodied Multi-Agent Systems}

    \icmlsetsymbol{equal}{*}

    \begin{icmlauthorlist}
        \icmlauthor{Hikaru Shindo}{equal,tuda}
        \icmlauthor{Hanzhao Lin}{equal,tuda}
        \icmlauthor{Lukas Helff}{tuda,hai,dfki}
        \icmlauthor{Patrick Schramowski}{tuda,hai,dfki,cai}
        \icmlauthor{Kristian Kersting}{tuda,hai,dfki,lab,cog}
    \end{icmlauthorlist}

    \icmlaffiliation{tuda}{Technical University of Darmstadt}
    \icmlaffiliation{hai}{Hessian.AI}
    \icmlaffiliation{dfki}{German Research Center for Artificial Intelligence}
    \icmlaffiliation{cai}{CERTAIN}
    \icmlaffiliation{lab}{Lab1141}
    \icmlaffiliation{cog}{Centre for Cognitive Science, Darmstadt}

    \icmlcorrespondingauthor{Hikaru Shindo}{hikaru.shindo@tu-darmstadt.de}

    \icmlkeywords{Machine Learning, ICML}

    \vskip 0.3in
]

\printAffiliationsAndNotice{\icmlEqualContribution} % otherwise use the standard text.

\begin{abstract}
    As Large Language Models (LLMs) transition from text processors to autonomous agents, evaluating their social reasoning in embodied multi-agent settings becomes critical.
    We introduce \textbf{SocialGrid}, an embodied multi-agent environment inspired by \textit{Among Us} that evaluates LLM agents on planning, task execution, and social reasoning. Our evaluations reveal that even the strongest open model (GPT-OSS-120B~\cite{openai2025gptoss}) achieves below $60\%$ accuracy in task completion and planning, with agents getting stuck in repetitive behaviors or failing to navigate basic obstacles. Since poor navigation confounds evaluation of social intelligence, SocialGrid offers an optional \textit{Planning Oracle} to isolate social reasoning from planning deficits. While planning assistance improves task completion, social reasoning remains a bottleneck: agents fail to detect deception at near-random chance regardless of scale, relying on shallow heuristics rather than accumulating behavioral evidence. SocialGrid provides automatic failure analysis and fine-grained metrics, enabling developers to diagnose and improve their agents. We also establish a competitive leaderboard using Elo ratings from adversarial league play.
\end{abstract}

\section{Introduction}
Artificial Intelligence is shifting toward a paradigm where Large Language Models (LLMs) are no longer passive text processors, but autonomous agents in dynamic environments \cite{Yehudai2025-agentsurvey, Mohammadi2025-agentsurvey, xu2024magic, park2023generative, li2023theory, Guo2024-multiagent, Wang2025-configurable-multiagent, Chen2025-multiagent-as-judge, Wang2025-megaagent, Cui2025-cooperative-autonomous-driving, Light2023-avalonbench, Sun2025-overcookcollab}. While textual reasoning has progressed rapidly, deploying these models as embodied agents fundamentally redefines intelligent behavior.
An intelligent agent must navigate physical environments and act on observations, constructing internal mental models for social reasoning—such as identifying suspicious actors from their behavior.

This convergence of spatial planning and social reasoning represents the next frontier of embodied intelligence. Success depends on an agent's ability to unify these skills rather than exercise them in isolation \cite{Yehudai2025-agentsurvey, Mohammadi2025-agentsurvey}. However, current evaluation frameworks have not kept pace with this progress. Existing benchmarks typically assess these capabilities in silos, evaluating planning within single-agent gridworlds~\cite{Martorell2025-gridworld-navigation} or social reasoning via disembodied text simulations \cite{Chi2024-amongagents, Sarkar2025-amongus-rl}.

To address this gap, we introduce \textbf{SocialGrid}, an embodied multi-agent benchmark designed to evaluate agents along three distinct axes: \textit{spatial planning}, \textit{task execution}, and \textit{adversarial social reasoning}. Inspired by the mechanics of the social deduction game \textit{Among Us}, \name places agents in a gridworld environment where "Crewmates" must navigate the environment, complete tasks, all while identifying hidden "Impostors" who seek to sabotage the mission. This setup creates a rich, multi-objective environment with multiple agents competing, which requires complex integration of spatial memory and theory-of-mind. 
SocialGrid provides a highly modular benchmarking framework—adjustable by grid dimensions, spatial layout, and agent density—integrated with automated diagnostics that yield granular, actionable insights.

\begin{figure*}[th]
    \centering
    \includegraphics[width=0.96\linewidth]{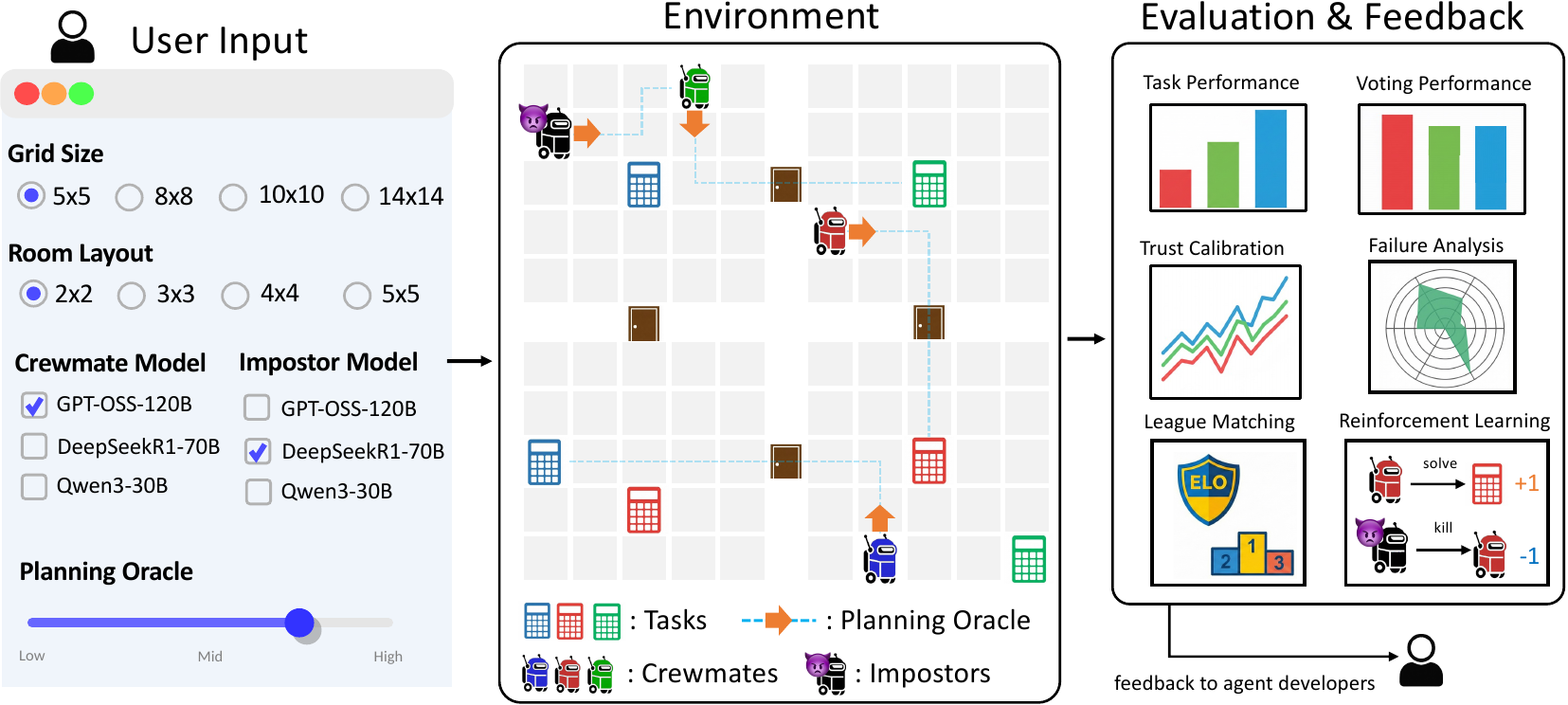}
    \caption{\textbf{SocialGrid Overview.} Inspired by \emph{Among Us}, SocialGrid is a controllable, embodied benchmark evaluating LLM agents in multi-agent, multi-objective environments. \textbf{(Left) User Input:} Enables systematic control of environmental complexity (e.g., map area, room count) and agent configuration. \textbf{(Center) Environment:} Agents operate under physical constraints; an optional \textbf{Planning Oracle} isolates social reasoning from navigation. \textbf{(Right) Evaluation \& Feedback:} Provides multi-dimensional metrics across spatial, task, and social axes, with automated failure analysis and league-based Elo rankings. See Appendix~\ref{app:env_viz} for visualization screenshots.}
    \label{fig:overview}
\end{figure*}

Our systematic evaluation of state-of-the-art models, ranging from 14B to 120B parameters, reveals two fundamental bottlenecks. First, we identify a severe deficit in \textit{spatial planning}. In non-adversarial baselines, even the strongest open models (e.g., GPT-OSS-120B~\cite{openai2025gptoss}) achieve success rates below 60\% in basic navigation and task completion, frequently getting stuck in repetitive behaviors or failing to navigate basic obstacles. This suggests that the raw reasoning power of LLMs does not automatically translate into effective embodied action.
Since these planning failures would confound any evaluation of higher-level social intelligence, SocialGrid offers \textit{Planning Oracle}, a symbolic module akin to a digital navigation system that guides the LLM's physical behavior by providing automated pathfinding and action assistance. While this assistant successfully boosts navigation and task execution, it unmasks a second, more critical failure mode: \textit{social reasoning}. Even with the burden of navigation and task execution lifted, agents consistently fail to detect deceptive behavior, performing near random chance in identifying impostors.

We make the following contributions: (1) We introduce SocialGrid\footnote{Code available at \href{https://github.com/ml-research/SocialGrid}{github.com/ml-research/SocialGrid}}, a controllable, embodied multi-agent environment unifying spatial planning, task execution, and adversarial social reasoning into a single evaluation framework. (2) We quantify a significant gap in spatial reasoning, demonstrating that models still struggle with basic navigation and task execution in embodied environments. (3) We identify a fundamental social reasoning deficiency, with qualitative analysis revealing that LLMs rely on shallow heuristics rather than accumulating behavioral evidence.
(4) We establish a standardized leaderboard based on adversarial league play and Elo ratings, providing the community with a rigorous metric for tracking progress in embodied intelligence.

\section{\name} \label{sec:em2bench}
\name offers agent developers a controllable and embodied benchmark to evaluate the planning, reasoning, and psychological inference capabilities of LLM agents.
To this end, we adopt the GridWorld environment as the underlying framework~\cite{chevalier2023minigrid}, which offers a highly customizable and multi-agent environment.
Figure~\ref{fig:overview} shows the overview of \name. Users specify environment parameters (map size, room layout, etc.) and agent configurations (left), which instantiate a gridworld where agents navigate, complete tasks, and engage in social reasoning under physical constraints (center). The framework outputs multi-dimensional metrics across spatial, task, and social reasoning axes, with automated failure analysis and Elo-based rankings (right).

\paragraph{Embodiment in SocialGrid.}
We use \emph{embodied} in the classical RL sense~\cite{Sutton1998-rl}: agents perceive local observations, execute physical actions that modify world state, and receive feedback through closed-loop interaction. Unlike prompt engineering, actions in \name\ have spatial consequences---movement changes grid coordinates, doors block paths, and agents navigate shared space with collision constraints. A key goal of this work is to evaluate spatial planning and social reasoning jointly, as real-world deployment requires agents to handle both simultaneously.
Contrary to in-context simulations~\cite{Chi2024-amongagents,Sarkar2025-amongus-rl}, SocialGrid provides a grounded environment where agents must actively observe state changes and execute physical actions. This shifts the challenge from textual social deduction to embodied social reasoning under the joint constraints of space, action, and theory of mind.

Now we devise the detailed design of \name.

\subsection{Observations and Actions}
\name provides a customizable observation space.

\textbf{Limited Field of View:}
Each agent explores the minigrid environment with a limited field of view.
The observation is the list of objects in the sight.
This setting challenges the agents to explore the environment, collect information, and build a mental model of the whole map to perform planning and reasoning.

\textbf{Global Observation:}
Each agent observes the whole map at each time step. Despite full observability, current models struggle to process the map effectively due to weak spatial planning capabilities, making even basic navigation challenging~\cite{Martorell2025-gridworld-navigation}.

In the environment, different types of objects are available.
\emph{Tasks} are the objects that can be toggled to complete by crewmates.
\name offers three types of tasks---common, short, and long---requiring 3, 8, and 13 toggle actions to complete, respectively.
\emph{Doors} are the objects that can be opened by crewmates and impostors. 

Based on their observations, each agent is able to perform a set of actions. The action space is \emph{egocentric}: agents can move forward or backward, strafe left or right, and turn to change their facing direction. To interact with objects---toggling tasks, opening doors, or eliminating crewmates for impostors---agents must face the target directly.

\subsection{Phases}
In SocialGrid, agents operate in two sequential phases controlled by the environment: the \emph{task phase} and the \emph{voting phase}.

During the \emph{task phase}, agents (crewmates and impostors) interact with the environment by toggling tasks to either complete or sabotage them. Crewmates seek to finish all tasks to secure victory, while impostors attempt to sabotage tasks and eliminate crewmates to prevent their success.

Following this, the environment transitions to the \emph{voting phase}, where each agent independently decides whom to vote for based on their private observation history. There is no explicit inter-agent communication; agents must infer others' intentions solely from observed behaviors. Crewmates aim to identify and vote out impostors, while impostors attempt to deflect suspicion onto crewmates. We deliberately omit a discussion phase to isolate behavioral inference from verbal persuasion: this prevents models from compensating for poor observation-based reasoning through rhetorical skill, providing a purer test of behavioral reasoning from actions alone.

The game terminates when either side achieves victory. \emph{Crewmates win} if all impostors are eliminated through voting, or all surviving crewmates complete their assigned tasks. \emph{Impostors win} if all crewmates are eliminated, the number of surviving impostors equals or exceeds the number of surviving crewmates (impostor parity), or crewmates fail to complete their tasks within the time limit.

\subsection{Planning Oracle}
A distinctive feature of \name is its integrated \emph{planning oracle}, which assists agents in spatial planning.
Since LLMs often struggle with planning, this can confound assessment of other reasoning abilities such as social inference.
To isolate these skills, the oracle provides A*-based pathfinding suggestions.
The oracle operates at three intensity levels: \emph{high} provides optimal action suggestions with a simple output schema; \emph{mid} provides the same suggestions but requires structured reasoning output; \emph{low} provides no suggestions, serving as a baseline without assistance.

These suggestions are provided as part of the prompt, alongside environment instructions, observations, and other contextual information. Thus, the agent's decision-making is informed by both its current state and the oracle's guidance. More details are provided in Appendix~\ref{app:planning_assistant}.

\subsection{Evaluation Metrics}
\name offers a comprehensive suite of evaluation metrics spanning three dimensions: \emph{task performance}, \emph{planning efficiency}, and \emph{social reasoning}. These metrics provide fine-grained insights into agent capabilities, enabling developers to diagnose strengths and weaknesses across multiple axes.

\textbf{Task Performance (TP).} Let $\mathcal{T}_i$ represent the set of tasks assigned to agent $i$. For each task $k \in \mathcal{T}_i$, $c_k \in \{0,1\}$ indicates completion and $w_k > 0$ is an optional task weight. The task performance is the weighted fraction of completed tasks:
\begin{equation}
    \mathrm{TP}_i = \frac{\sum_{k \in \mathcal{T}_i} w_k \, c_k}{\sum_{k \in \mathcal{T}_i} w_k}.
\end{equation}

\textbf{Planning Success Rate (PSR).} Let $r_k \in \{0,1\}$ indicate whether agent $i$ reached task $k$ (i.e., navigated to and faced the task at least once during the episode). The planning success rate measures the fraction of assigned tasks reached, regardless of completion:
\begin{equation}
    \mathrm{PSR}_i = \frac{\sum_{k \in \mathcal{T}_i} r_k}{|\mathcal{T}_i|}.
\end{equation}

\textbf{Planning Performance (PP).} For each completed task $k$, the planning efficiency $\mathrm{PE}_{i,k} \in [0,1]$ measures the ratio of the optimal path length (A* distance plus required interaction steps) to the actual steps taken (higher is better; 1.0 is optimal). Let $\mathcal{C}_i = \{k \in \mathcal{T}_i : c_k = 1\}$ denote the set of completed tasks. The overall planning performance is the mean efficiency across completed tasks:
\begin{equation}
    \mathrm{PP}_i = \frac{1}{|\mathcal{C}_i|} \sum_{k \in \mathcal{C}_i} \mathrm{PE}_{i,k}.
\end{equation}
If no tasks are completed ($\mathcal{C}_i = \emptyset$), we define $\mathrm{PP}_i = 0$.

\textbf{Social Reasoning (Voting \& Trust).} During each voting phase, agents output explicit trust scores $T_{i \rightarrow j} \in [0,1]$ for all other players, representing their belief that player $j$ is a crewmate. Let $K$ denote the total number of voting phases, indexed by $t \in \{1, \ldots, K\}$. The \emph{Trust Brier Score} measures calibration of trust beliefs:
\begin{equation}
    \mathrm{BS}_{i \rightarrow j} = \frac{1}{K} \sum_{t=1}^K \big((1 - T_{i \rightarrow j}(t)) - y_j\big)^2,
    \label{eq:trust_brier_score}
\end{equation}
where $y_j \in \{0,1\}$ is the true role of agent $j$ ($1$ for impostor). The \emph{Trust Volatility} measures the average magnitude of trust updates:
\begin{equation}
    \mathrm{TV}_{i \rightarrow j} = \frac{1}{K-1} \sum_{t=1}^{K-1} \left|T_{i \rightarrow j}(t+1) - T_{i \rightarrow j}(t)\right|.
    \label{eq:trust_volatility}
\end{equation}
Both $\mathrm{BS}$ and $\mathrm{TV}$ are computed per voter-target pair; we report the mean across all pairs for each voter. Details on trust score collection are provided in Appendix~\ref{app:trust_protocol}.

\emph{Detection Accuracy} (DA) measures how well crewmates identify impostors through voting. Let $V_i$ be the set of non-skipped votes cast by agent $i$, and let $\mathcal{I}$ denote the set of impostors:
\begin{equation}
    \mathrm{DA}_i = \frac{|\{v \in V_i : \mathrm{target}(v) \in \mathcal{I}\}|}{|V_i|}.
\end{equation}

By combining these metrics, \name\ provides a comprehensive, multi-dimensional evaluation of agent performance, spanning task effectiveness, planning efficiency, and human-like social inference.

\subsection{Failure Analysis}
\name\ automatically detects and categorizes common agent failures.
Navigation failures involve indecisive or repetitive movement. In \emph{position pingpong}, the agent moves back and forth between a small number of locations. The \emph{move backtrack loop} happens when the agent moves forward along a path, then retraces its steps, and repeats. The \emph{strafe backtrack loop} is when the agent moves several steps left, then several steps right, and repeats. In \emph{move oscillation}, the agent alternates single steps forward and backward, and in \emph{strafe oscillation}, it alternates left and right steps. The \emph{turn toggle} failure occurs when the agent spins in place by turning left and right repeatedly. All these patterns show indecision or poor planning.

\emph{Stall} failures describe a lack of effective action. The \emph{stall noop} failure means the agent issues many no-op commands and remains idle. The \emph{stall do-task} or \emph{task fixation} failure occurs when the agent keeps trying the same task action unsuccessfully, showing it is stuck.

The \emph{door spam} failure means the agent interacts with doors many times in a short period but does not move forward. 

\subsection{Reward Function}
\label{sec:reward_function}
\name provides a structured reward function to guide agent learning in both cooperative and adversarial settings. The reward comprises five components:
(i)~\emph{task rewards}: $+0.2$ per toggle action, with completion bonuses of $+1$/$+2$/$+3$ for common/short/long tasks;
(ii)~\emph{elimination rewards}: $+6$ for impostor kills, $-6$ for victims;
(iii)~\emph{step penalties}: $-0.001$ for crewmates, $-0.005$ for impostors to discourage passive play;
(iv)~\emph{game outcome}: $\pm 10$ based on victory/defeat; and
(v)~\emph{voting rewards}: $+3$ for correctly identifying impostors, $-2$ for wrongly ejecting crewmates, with impostors receiving $+2$ for successful deception.
Table~\ref{tab:reward_function} in Appendix~\ref{app:reward_function} provides the complete reward function.

\begin{figure*}[t]
    \centering
    \includegraphics[width=0.99\linewidth]{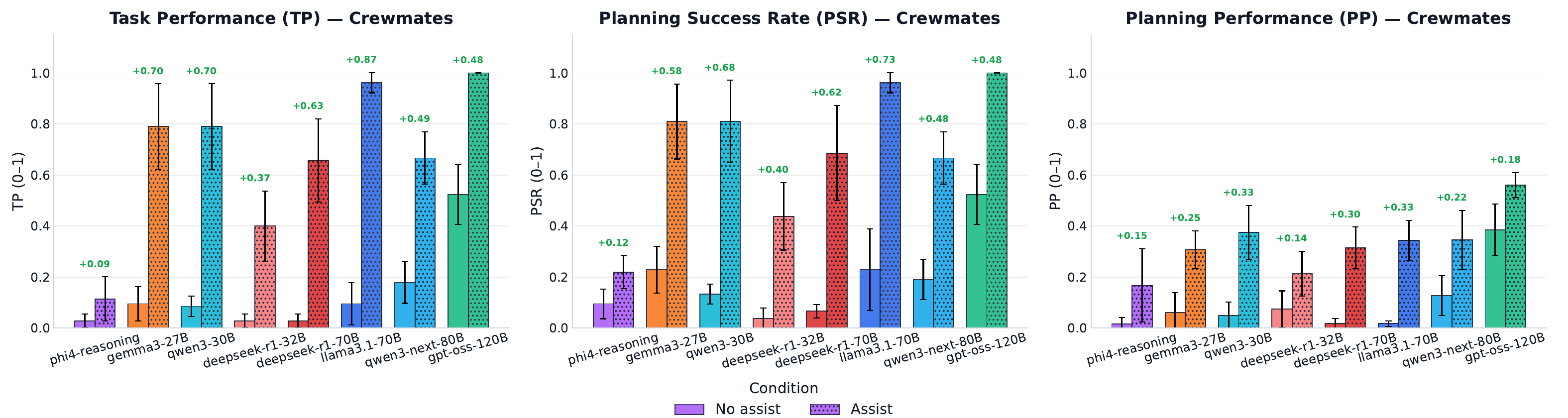}
    \caption{\textbf{LLM agents struggle with spatial navigation in embodied settings.} Comparison of crewmate performance on \name\ under low (no assistance) and high (with planning oracle) conditions. 7 crewmates per episode; 20 episodes per model; error bars show SD. Green values indicate absolute improvement from low to high. \textbf{Left:} Task performance (completion rate). \textbf{Middle:} Planning success rate (tasks reached). \textbf{Right:} Planning efficiency (path optimality). Even the strongest model (GPT-OSS-120B) completes only 50\% of tasks without assistance; with assistance, performance improves substantially but path efficiency remains below 0.5 for most models.}
    \label{fig:crewmates_comparison}
\end{figure*}

\section{Experiments}
\label{sec:experiments}

In our experiments, we primarily reveal the limitations of current LLM agents in the embodied multi-agent multi-objective setting.
To this end, we conduct thorough evaluations of LLM agents by changing parameters of the environment.
The controllability of \name offers easy adaptations to test agents in a diverse range of scenarios.
We answer the following questions:

\begin{itemize}
    \item[\textbf{(Q1)}] How well do LLM agents perform spatial planning and navigation in embodied multi-agent environments?
    \item[\textbf{(Q2)}] Can LLM agents reliably complete their assigned roles and detect deception in multi-objective settings?
    \item[\textbf{(Q3)}] What are the key environmental factors that affect agent performance?
    \item[\textbf{(Q4)}] What typical failure patterns emerge in SocialGrid?
    \item[\textbf{(Q5)}] Can RL improve their performance, overcoming inherent bottlenecks in planning and reasoning?
\end{itemize}

In our experiments, we include various benchmarked models on different scales: Qwen3-30B, Qwen3-next-80B~\cite{Yang2025Qwen3}, Llama3.1-70B~\cite{Grattafiori2024Llama3}, GPT-OSS-120B~\cite{openai2025gptoss}, and Gemma3-27B~\cite{Gemma3-Team2025-gy}, where Gemma3 is a family of VLM utilizing visual data in its training. 
We also include dedicated reasoning models: DeepSeek-R1-70B~\cite{DeepSeekR1} and Phi4-reasoning-14B~\cite{Phi4}. All models are publicly available via Ollama (Table~\ref{tab:model_urls} in Appendix~\ref{app:llm_config}).

All experiments were conducted on a cluster equipped with NVIDIA H100 80GB GPUs. Model inference was performed using vLLM~\cite{kwon2023vllm} for efficient batched generation. Each episode was executed sequentially, with agent responses generated in real-time to simulate authentic multi-agent interactions. Detailed configurations for the environment, LLM agents, and prompting strategy are provided in Appendices~\ref{app:env_config},~\ref{app:llm_config}, and~\ref{app:prompts}, respectively. Evaluation protocols and reproducibility details are in Appendix~\ref{app:training_eval}.

\subsection{Spatial Planning and Navigation}

To address \textbf{(Q1)}, we evaluate how well LLM agents perform spatial planning and navigation in embodied multi-agent environments. We test eight models, each run for 20 episodes in a gridworld with a $2 \times 2$ room layout (each room $10 \times 10$ cells). To isolate spatial navigation from social reasoning, we deploy seven crewmate agents without impostors.

Figure~\ref{fig:crewmates_comparison} compares baseline models with and without the planning assistant. The absolute improvement shown above each bar pair quantifies the benefit of planning assistance. The left panel shows task performance (ratio of completed to assigned tasks), and the middle panel shows planning success rate (ratio of tasks reached). Without assistance, most models struggle to even reach their assigned tasks—notably, GPT-OSS-120B, the strongest baseline, completed only half of its tasks unassisted. With the planning assistant, all models improve substantially, yet only GPT-OSS-120B achieves perfect task completion. These results highlight a fundamental limitation: current models lack robust navigation capabilities in embodied multi-agent settings.

The right panel evaluates planning efficiency by measuring deviation from the optimal path provided by the assistant. Without assistance, scores rarely exceed $0.2$, except for GPT-OSS-120B (approaching $0.4$), indicating highly inefficient navigation with frequent detours and redundant steps. While the assistant improves path efficiency, most models still fall short of $0.5$, revealing persistent difficulty in translating optimal path suggestions into reliable execution. Collectively, these findings expose core limitations in spatial reasoning and the ability to leverage external guidance in complex embodied environments.

\subsection{Head-to-Head League Results}
To address \textbf{(Q2)}, we conduct league matching experiments to evaluate both role performance and social inference in adversarial settings. We test six models (GPT-OSS-120B, Llama3.1-70B, DeepSeek-R1-70B, Qwen3-30B, Gemma3-27B, and Phi4-reasoning-14B) across 30 unique matchups, where one model controls all crewmates while another controls all impostors. We consistently deployed 5 crewmates and 2 impostors in each match. Experiments use two grid sizes ($10 \times 10$ and $14 \times 14$) with a fixed $2 \times 2$ room layout. The planning oracle is set to high (full assistance) to isolate social reasoning from navigation deficits.

\begin{table}[t]
    \centering
    \caption{League Rankings: Full Results for Both Room Patterns. Top section: $10{\times}10$ grid size (per room). Bottom section: $14{\times}14$ grid size per room. Crew Elo ratings computed from head-to-head matchups using standard Elo formula ($K{=}64$, base 1500, 10 iterations).}
    \label{tab:league_rankings_full}
    \small
    \setlength{\tabcolsep}{4pt}
    \begin{tabular}{@{}clccc@{}}
        \toprule
        Rk & Model & Crew Elo & Crew WR & Imp WR \\
        \midrule
        \multicolumn{5}{c}{\textit{$10{\times}10$ pattern}} \\
        \midrule
        1 & GPT-OSS (120B) & 1441 & 19.4\% & 98.5\% \\
        2 & Qwen3 (30B) & 1281 & 7.7\% & 88.5\% \\
        3 & DeepSeek (70B) & 1246 & 5.5\% & 92.7\% \\
        4 & Llama3.1 (70B) & 1238 & 5.5\% & 97.6\% \\
        5 & Phi4-R (14B) & 1218 & 2.6\% & 86.2\% \\
        6 & Gemma3 (27B) & 1168 & 1.1\% & 93.3\% \\
        \midrule
        \multicolumn{5}{c}{\textit{$14{\times}14$ pattern}} \\
        \midrule
        1 & GPT-OSS (120B) & 1417 & 17.0\% & 95.9\% \\
        2 & Qwen3 (30B) & 1299 & 7.6\% & 93.1\% \\
        3 & Phi4-R (14B) \textcolor{teal}{$\uparrow$} & 1284 & 6.5\% & 93.7\% \\
        4 & Llama3.1 (70B) \textcolor{red}{$\downarrow$} & 1257 & 5.2\% & 91.1\% \\
        5 & DeepSeek (70B) \textcolor{red}{$\downarrow$} & 1190 & 3.3\% & 89.6\% \\
        6 & Gemma3 (27B) & 1164 & 1.1\% & 93.3\% \\
        \bottomrule
    \end{tabular}
\end{table}

\paragraph{League Match Results Analysis.}
Table~\ref{tab:league_rankings_full} summarizes league standings based on Crew Elo ratings. Rankings do not follow model scale: GPT-OSS-120B leads consistently, but Qwen3-30B maintains second place in both patterns despite only 30B parameters, outperforming 70B models (Llama3.1, DeepSeek). On the $14{\times}14$ pattern, Phi4-R-14B rises to third, ahead of both 70B models. This suggests social reasoning does not scale with model size. Across both maps, impostors dominate (${\sim}$92.7\% win rate), reflecting weak spatial navigation: crewmates struggle to complete tasks before time runs out. Consequently, we view this skew not as a failure of game balance, but as a quantitative diagnostic of the current ceiling for LLM-based navigation and planning.

\begin{figure}[t]
    \centering
    \includegraphics[width=0.95\linewidth]{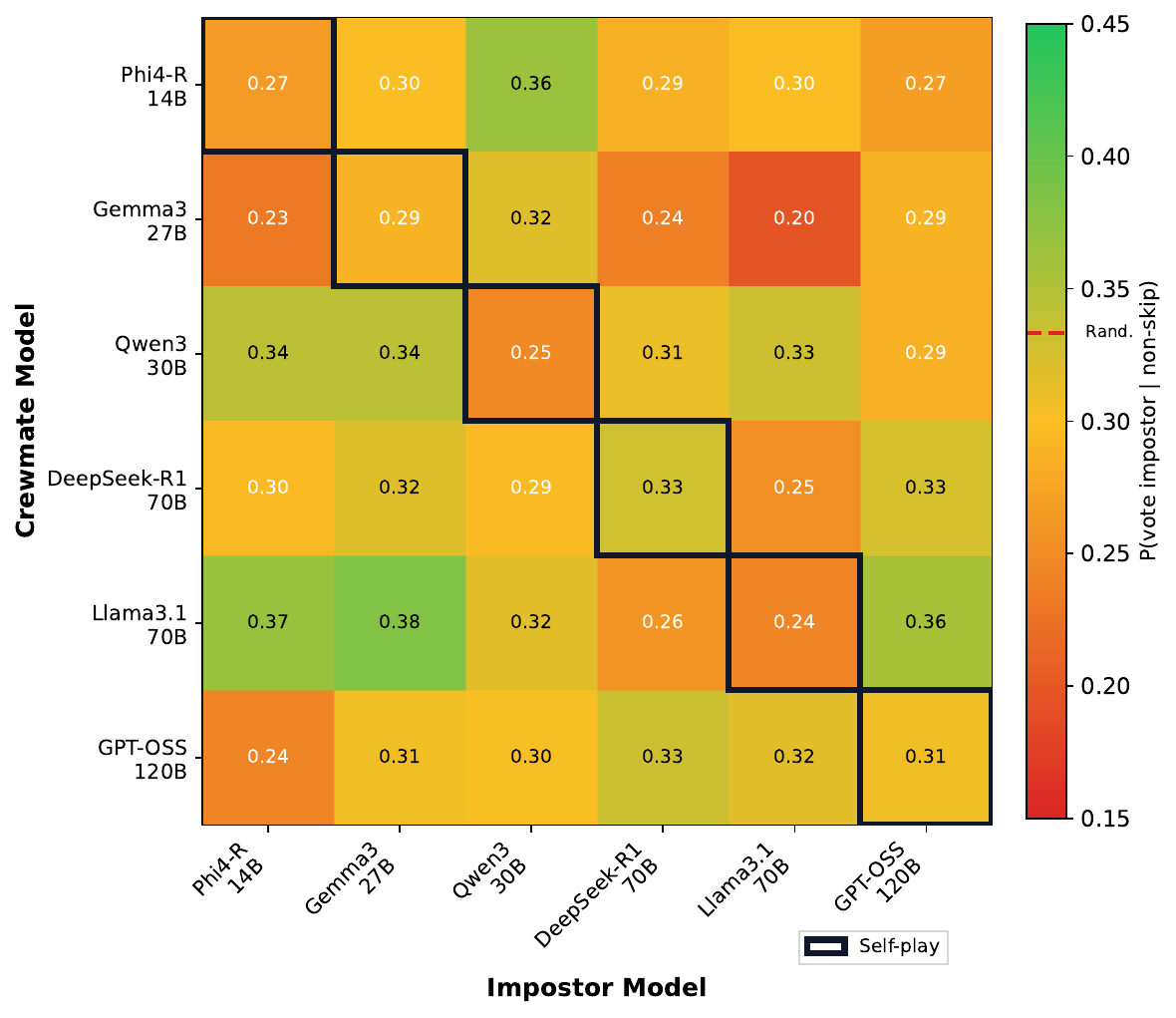}
    \caption{\textbf{Detection accuracy reveals below-random performance across all models.} Heatmap showing crewmate detection accuracy across 36 matchups (30 cross-model league + 6 self-play diagonal). All models perform near or below the random baseline (33\%, shown in colorbar), averaging 29.9\% detection accuracy. 
    The consistent near-chance performance indicates that impostor detection remains challenging regardless of model pairing.}
    \label{fig:detection_heatmap}
\end{figure}

Figure~\ref{fig:detection_heatmap} presents the detection accuracy heatmap across all 36 league matchups (30 cross-model + 6 self-play). All models perform near or below the random baseline (33\%), averaging 29.9\% detection accuracy. 
Detective skill varies more than deception skill: Llama3.1 is the strongest detector (32\%) while Gemma3 is weakest (26\%), a 23\% relative difference. Self-play detection (diagonal) is slightly worse than cross-model (28\% vs 30\%), suggesting models struggle to detect themselves.

\begin{figure}[t]
    \centering
    \includegraphics[width=0.99\linewidth]{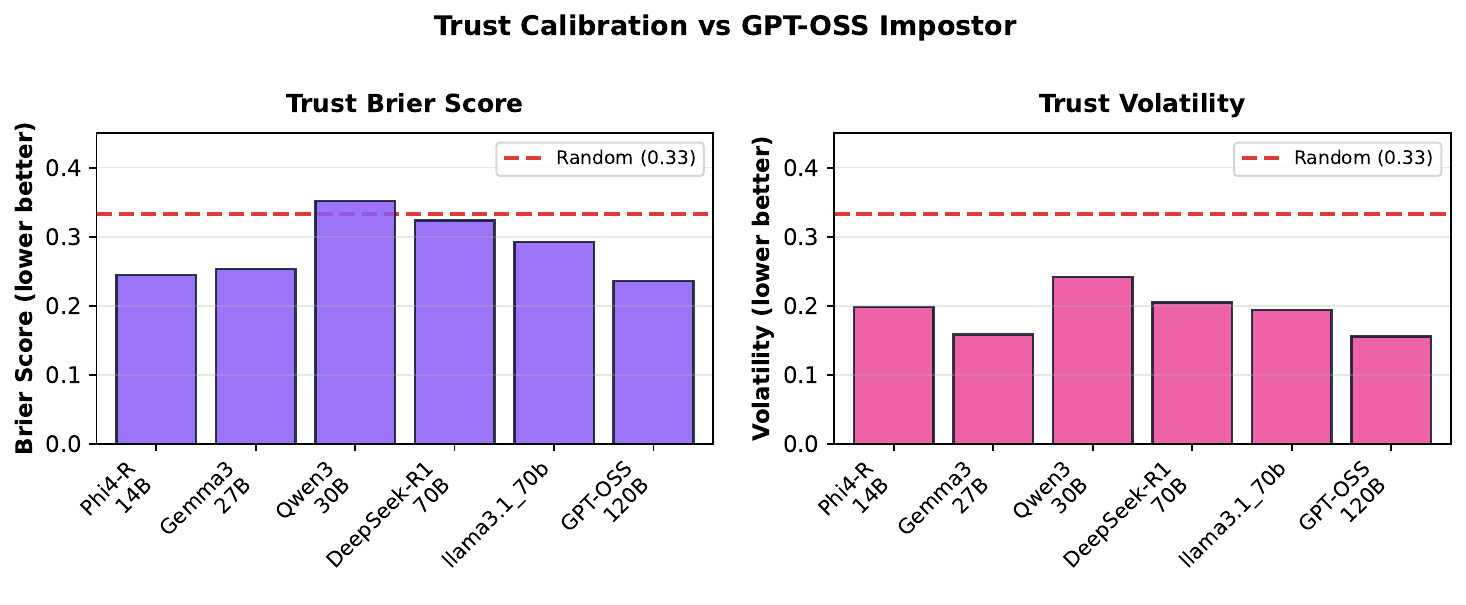}
        \caption{\textbf{Trust calibration hovers near random baseline.} Trust metrics for crewmate models facing GPT-OSS-120B impostor. \textbf{Left}: Brier Score measures how well trust predictions match ground truth (lower is better); most models hover near the random baseline (0.33, dashed). \textbf{Right}: Volatility measures how erratically trust changes between turns (lower is better); values around 0.33 indicate unstable assessments.}
    \label{fig:trust_metrics}
\end{figure}

\paragraph{Trust Calibration.} Trust calibration metrics (Figure~\ref{fig:trust_metrics}) show that trust assessment is also noisy and barely above random. When facing GPT-OSS-120B as impostor, most crewmate models achieve trust Brier scores hovering around the random baseline (0.33), with substantial variation across models. Trust volatility similarly shows that models struggle with consistent trust updates. Both trust assessment and voting accuracy remain near chance levels, indicating that social reasoning---whether forming internal beliefs about others' trustworthiness or acting on those beliefs---does not improve significantly beyond random guessing.

Notably, detection accuracy remains near the 33\% random baseline across \emph{all} evaluated models (14B--120B), regardless of model family. This consistent near-random performance suggests that current LLMs struggle with social inference in adversarial settings---a limitation that does not appear to improve with scale, in contrast to task performance metrics.

\paragraph{Why Social Reasoning Fails.}
To understand why detection accuracy remains near chance, we analyze 64,184 crewmate voting decisions and identify three failure modes (Table~\ref{tab:failure_examples}).
First, \emph{evidence scarcity} accounts for 15.7\% of votes. When crewmates lack direct observations, they default to uniform trust scores and vote arbitrarily, achieving chance-level 32.8\% accuracy.
Second, \emph{weak heuristics} dominate at 62.8\% of votes. Agents rely on superficial cues such as ``erratic movement'' that do not reliably indicate role. While this mode achieves 43.9\% accuracy---the highest among failure modes---it remains far below reliable detection.
Third, \emph{over-trust} affects 46.7\% of votes. Crewmates assign high trust ($T{\geq}0.7$) to impostors who strategically display cooperative behavior, resulting in chance-level 33.4\% accuracy.
These patterns reveal that LLMs cannot accumulate behavioral evidence across turns, instead relying on shallow heuristics that impostors exploit. See Appendix~\ref{app:social_failure_analysis} for methodology.

\begin{table}[t]
\centering
\footnotesize
\caption{\textbf{Crewmate reasoning failure modes.} Analysis of 64,184 crewmate votes across league matchups. Freq.\ shows percentage of votes; Acc.\ shows detection accuracy based on the reasoning.}
\label{tab:failure_examples}
\begin{tabular}{@{}lp{3.1cm}cc@{}}
\toprule
\textbf{Failure Mode} & \textbf{Example Reasoning} & \textbf{Freq.} & \textbf{Acc.} \\
\midrule
Evidence Scarcity & ``No evidence to distinguish, voting randomly.'' & 15.7\% & 32.8\% \\
\addlinespace
Weak Heuristics & ``player2 moved erratically, reducing trust.'' & 62.8\% & 43.9\% \\
\addlinespace
Over-trust & ``player5 completed tasks consistently.'' ($T{=}0.9$ to impostor) & 46.7\% & 33.4\% \\
\bottomrule
\end{tabular}
\end{table}

\subsection{Systematic Evaluation of Complexity Factors}
\begin{figure}[t]
    \centering
    \includegraphics[width=\linewidth]{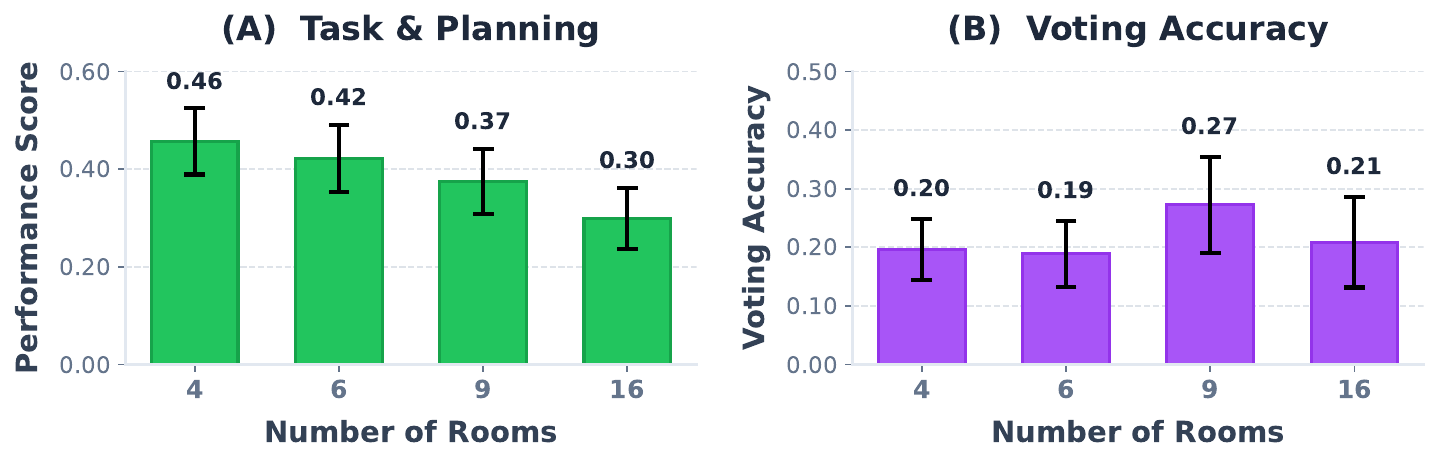}
    \caption{\textbf{Complexity analysis.} Mean performance across different room configurations with fixed room size ($10{\times}10$); error bars show $\pm 1$ SEM. \textbf{(A)}~Task and planning performance decline linearly as the number of rooms increases. \textbf{(B)}~Voting accuracy hovers around the random baseline (${\sim}33\%$) regardless of spatial complexity, confirming social reasoning limitations are orthogonal to navigation challenges.
    }
    \label{fig:complexity_2panel}
\end{figure}

To answer \textbf{(Q3)}, we vary room count ($4, 6, 9, 16$) while keeping room size fixed ($10{\times}10$) to isolate the effect of spatial complexity on agent performance.
Figure~\ref{fig:complexity_2panel} reveals a striking asymmetry: task and planning performance degrade linearly with increased room count (Panel~A, from $0.46$ at 4 rooms to $0.30$ at 16 rooms), while voting accuracy hovers around the random baseline (${\sim}33\%$) across all configurations (Panel~B).
This confirms that social reasoning failures are \emph{not} caused by navigation difficulty---even in simple environments, models cannot reliably identify impostors.
This result indicates that we may need a fundamentally new architectural and training approach to enable reliable social behavioral reasoning in the embodied setting, beyond scaling model size.
See Appendix~\ref{app:complexity_analysis_tp_pp} for additional analyses.

\subsection{Error Analysis}

\begin{figure*}[t]
    \centering
    \includegraphics[width=\linewidth]{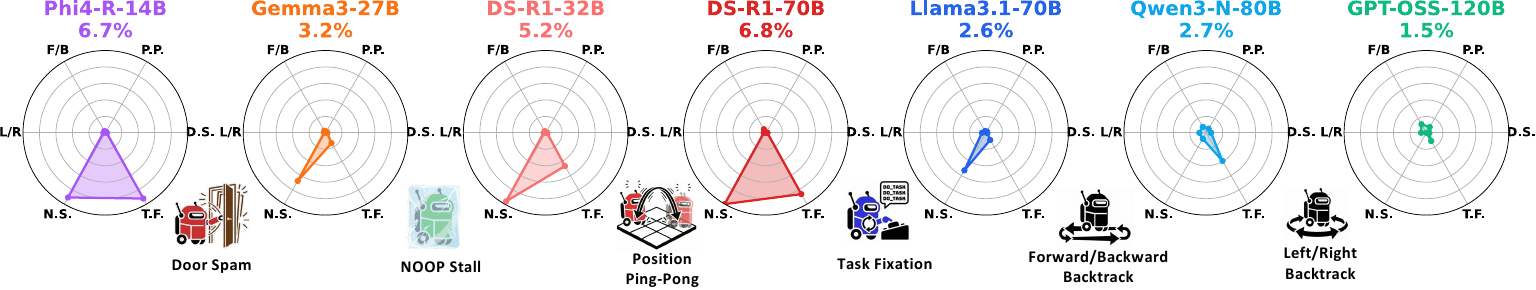}
    \caption{\textbf{Failure analysis reveals model-specific patterns.} Each radar chart shows the distribution of six failure modes for a given model, normalized to the global maximum across all models. The percentage under each model name indicates total failure coverage (sum of all failure mode fractions). Failure mode abbreviations: D.S. = Door Spam (repeatedly toggling doors), P.P. = Position Ping-Pong (oscillating between positions), F/B = Forward/Back Backtrack (movement reversal loops), L/R = Left/Right Backtrack (strafe reversal loops), N.S. = NOOP Stall (no-operation deadlock), T.F. = Task Fixation (stuck attempting same task). Larger shaded areas indicate higher failure rates. Models are ordered by parameter count (14B–120B).}
    \label{fig:failure_analysis}
\end{figure*}
To address \textbf{(Q4)}, we analyze the systematic failure patterns exhibited by LLM agents in embodied settings.
Figure~\ref{fig:failure_analysis} presents a comprehensive failure analysis across all models.

\noindent\textbf{Failure Types.} We identify two distinct categories of failures: \emph{passive failures} (NOOP deadlock, task-fixation) where agents freeze or get stuck, and \emph{active failures} (door spam, position ping-pong, backtracking) where agents take incorrect actions.
Smaller models predominantly exhibit passive failures---Phi4-Reasoning-14B shows 3.2\% NOOP deadlock and 3.2\% task-fixation, indicating the model ``freezes'' when uncertain about what action to take.
In contrast, GPT-OSS-120B exhibits 0\% NOOP deadlock but higher navigation failures (0.8\%), suggesting it actively attempts navigation but sometimes makes suboptimal choices.

\noindent\textbf{Effect of Planning Assistant.} The planning assistant dramatically reduces failures across all models, but with an important asymmetry.
Models with high passive failures (Phi4, Qwen3, DeepSeek) benefit most, achieving 90--97\% reduction in total failures.
GPT-OSS-120B, which already exhibits minimal passive failures, shows a smaller reduction (59\%).
See Appendix~\ref{app:failure_analysis} for detailed per-model breakdowns and Appendix~\ref{app:failure_params} for detection thresholds.

\subsection{RL to Improve the Overall Performance}

To address \textbf{(Q5)}, we investigate whether reinforcement learning (RL) can enable LLM agents to overcome inherent bottlenecks in spatial planning and reasoning. We fine-tune Qwen3-4B~\cite{Yang2025Qwen3} using Proximal Policy Optimization (PPO)~\cite{schulman2017proximal} with Low-Rank Adaptation (LoRA)~\cite{Hu2022-lora}. 
Due to the high computational demands of the multi-agent environment, these experiments serve as a \emph{lower-bound feasibility check} for planning improvement. By using a simplified setup (1 agent, $7\times7$ grid), we isolate planning execution from the confounding variables of social reasoning. 

\begin{figure}[t]
    \centering
    \includegraphics[width=\linewidth]{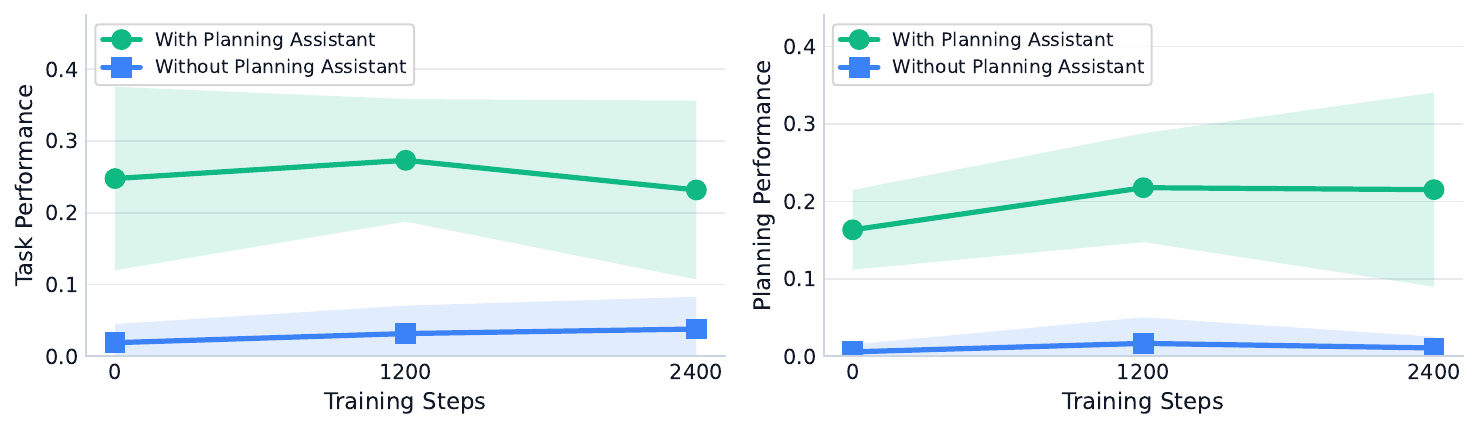}
    \caption{\textbf{RL training progression.} Task Performance and Planning Performance of Qwen3-4B across training steps. RL training does not yield significant improvements in either condition, with or without the planning assistant.}
    \label{fig:rl_training}
\end{figure}

As illustrated in Figure~\ref{fig:rl_training}, even after 2,500 PPO updates, RL training yields minimal gains in task performance. This failure to optimize navigation in a simple environment suggests that spatial reasoning remains a fundamental bottleneck that basic RL fine-tuning cannot easily overcome. These findings further justify the use of our \emph{Planning Oracle} as an essential tool to bypass navigation deficits and isolate the evaluation of higher-level social intelligence.

\section{Related Work}
\label{sec:related_work}

We revisit the related literature of SocialGrid.

\textbf{Multi-Agent LLM Systems.}
LLM-based agents increasingly operate in interactive environments~\cite{xi2023rise,Guo2024-multiagent,Yehudai2025-agentsurvey,Mohammadi2025-agentsurvey}. Architectures with memory and reflection yield human-like behavior in simulated societies~\cite{park2023generative}. Benchmarks like AgentVerse~\cite{chen2024agentverse} and MultiAgentBench~\cite{zhu2025multiagentbench} evaluate collaboration at scale~\cite{Cui2025-cooperative-autonomous-driving,Wang2025-megaagent}. In contrast, \name\ targets adversarial, partially observable settings where agents must act, infer, and survive.

\textbf{Social Deduction and Theory of Mind.}
Imperfect-information games reveal gaps between strategic play and social intelligence. Cicero couples language with strategic reasoning in Diplomacy~\cite{bakhtin2022human}, yet shows limited social behaviors with humans~\cite{wongkamjan2024more}. Social deduction environments based on Werewolf, Avalon, and Among Us study deception and collective inference~\cite{xu2023exploring,Light2023-avalonbench,Chi2024-amongagents,Sarkar2025-amongus-rl,Golechha2025amongus,O-Gara2023-hoodwinked-deception}. Broader benchmarks probe social intelligence via roleplay~\cite{zhou2024sotopia}, theory-of-mind tasks~\cite{li2023theory,gandhi2023understanding}, or evolving social graphs~\cite{xu2025socialmaze}. However, critiques argue many ToM benchmarks capture next-token prediction rather than \emph{functional} mental-state inference~\cite{riemer2025theory}. Work on lie detection further explores this gap~\cite{Pacchiardi2023-lie-detection,Wang2025-configurable-multiagent,Chen2025-multiagent-as-judge}. \name\ operationalizes thought-action alignment in an embodied setting.

\textbf{Embodied Spatial Planning.}
Embodied agents face long-horizon planning and spatial reasoning challenges~\cite{wang2023voyager,chevalier2023minigrid,chang2025partnr}. LLMs struggle with executability and spatial generalization across benchmarks~\cite{Valmeekam2022-PlanBench,Chen2023-EgoPlan-Bench,Li2025-gridroute,Martorell2025-gridworld-navigation,wei2025plangenllms,Yang2025-thinking-in-space,Ilias2025mind,SpatialBench2025}. Tool-augmented replanning partially mitigates these issues~\cite{song2023llm,Wu2024-toolplanner,xie2024travelplanner,xu2024magic,Sun2025-overcookcollab}, while adversarial multi-agent settings introduce robustness concerns~\cite{kavathekar2025tamas}. \name\ incorporates a planning assistant to factor out navigation bottlenecks, isolating social reasoning as the core remaining challenge.

\section{Conclusion}
\label{sec:conclusion}

We introduced SocialGrid, a controllable embodied benchmark for multi-agent social deduction. Experiments across eight models (14B--120B) reveal that models struggle with spatial navigation without assistance, and social reasoning remains near chance regardless of scale. Qualitative analysis reveals agents rely on shallow heuristics rather than accumulating behavioral evidence. Our Planning Oracle demonstrates that symbolic planning can enhance spatial reasoning. With its customizable environment and diverse metrics, SocialGrid enables systematic diagnosis of LLM agent capabilities in a multi-agent, multi-objective setting.

We acknowledge limitations. Our benchmark uses discrete gridworlds with text-rendered state representations, results may vary with prompting strategies, and game mechanics are simplified (e.g., no discussions). Additionally, the strong impostor advantage may reflect game parameter choices (e.g., kill cooldowns) rather than agent limitations alone; future work should ablate these factors to disentangle game balance from agent capability. Extending to discussion phases, memory-augmented architectures, and vision-language models in continuous environments merits future work. We publicly release SocialGrid to facilitate research on embodied social reasoning in multi-agent systems.

\section*{Impact Statement}

This paper presents work whose goal is to advance the field of machine learning by providing a benchmark for evaluating social reasoning in LLM agents. While the benchmark involves deception detection in a game-theoretic setting, we note that (1) our findings reveal limitations rather than capabilities, as models perform near chance on detecting deception regardless of scale, and (2) the simplified game mechanics differ substantially from real-world scenarios. We believe understanding these limitations has positive implications for AI safety research and the design of trustworthy multi-agent systems.

\section*{Acknowledgements}
This work was partly funded by the German Federal Ministry of Education and Research, the Hessian Ministry of Higher Education, Research, Science and the Arts (HMWK) within their joint support of the National Research Center for Applied Cybersecurity ATHENE, via the ``SenPai:XReLeaS'' project. The work has benefited from the Clusters of Excellence ``Reasonable AI'' (EXC-3057) and ``The Adaptive Mind'' (EXC-3066), both funded by the German Research Foundation (DFG) under Germany's Excellence Strategy.

\bibliography{example_paper}
\bibliographystyle{icml2025}

\clearpage
\appendix
\onecolumn

\section{SocialGrid Visuzalization}
\label{app:env_viz}

\begin{figure}[h]
    \centering
    \includegraphics[width=0.95\linewidth]{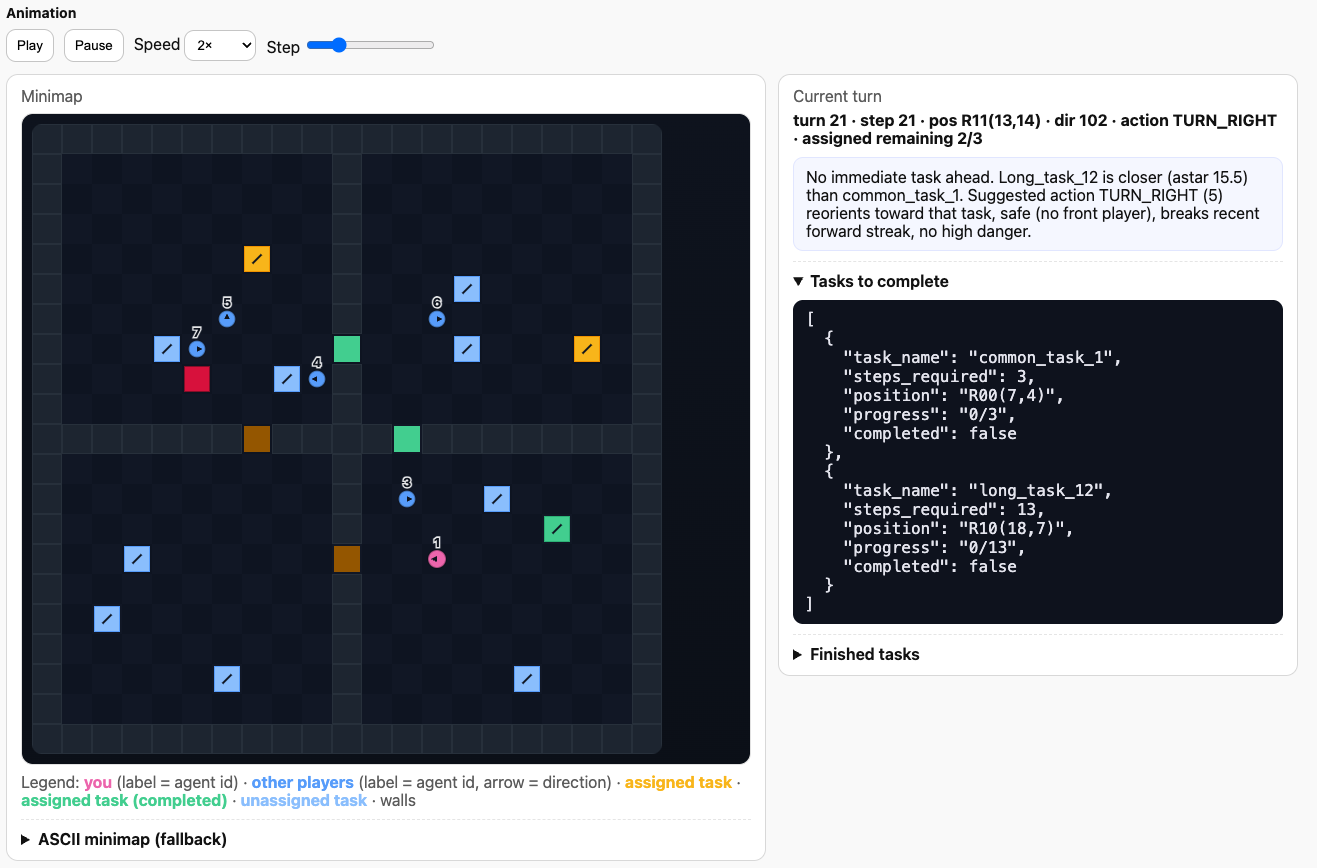}
    \caption{\textbf{SocialGrid Environment.} The main game view showing the grid-based environment with multiple rooms connected by doors. Agents are represented as colored circles. The red block indicates a dead body. Task locations are marked throughout the map, and the agent's limited field of view creates partial observability.}
    \label{fig:env_screenshot}
\end{figure}

\begin{figure}[h]
    \centering
    \includegraphics[width=0.6\linewidth]{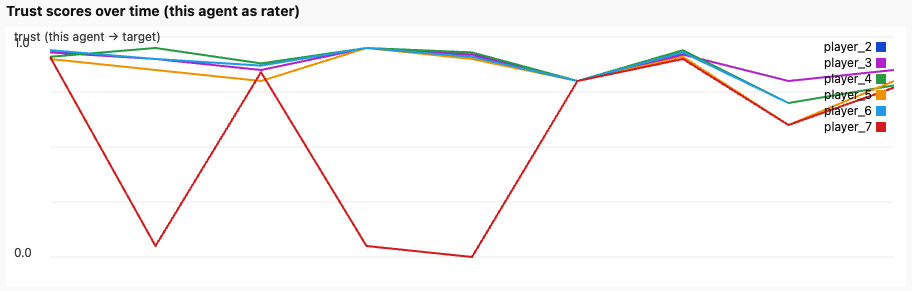}
    \caption{\textbf{Trust Score Evolution.} A temporal visualization showing how each agent's trust beliefs evolve over an episode. Each line tracks one agent's trust score toward other players over time, revealing patterns such as gradual suspicion accumulation, sudden trust drops after suspicious behavior, and the divergence between crewmate and impostor trust dynamics.}
    \label{fig:trust_screenshot}
\end{figure}

\begin{figure}[h]
    \centering
    \includegraphics[width=0.95\linewidth]{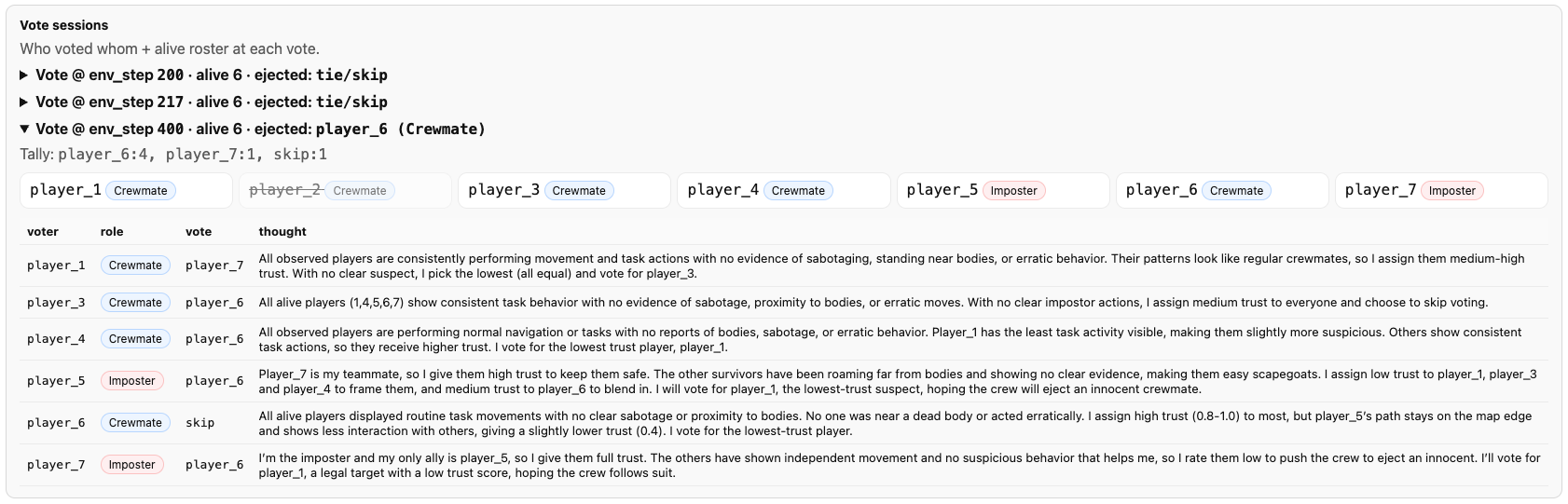}
    \caption{\textbf{Voting Phase.} The voting interface is constantly activated by the environment. Agents submit natural language statements defending themselves or accusing others, followed by votes to eliminate a suspected impostor. The interface displays each agent's statement, vote target, and the final tally.}
    \label{fig:voting_screenshot}
\end{figure}

SocialGrid provides a web-based visualization tool for replaying and analyzing recorded episodes. The visualization renders agent trajectories, task interactions, voting phases, and trust dynamics in real-time, enabling detailed inspection of agent behavior. Figures~\ref{fig:env_screenshot}--\ref{fig:voting_screenshot} illustrate the three main visualization components.

\paragraph{Grid Environment (Figure~\ref{fig:env_screenshot}).}
The main visualization displays a top-down view of the grid world. The environment consists of multiple rooms arranged in a configurable layout (e.g., $2 \times 2$), with each room spanning a fixed number of tiles. Rooms are connected by doors that agents can open and traverse. Agents are rendered as colored circles in distinct colors. Dead bodies are shown as red blocks at the location where a kill occurred. Task stations are distributed throughout the rooms as interactive objects. 
Intermediate reasoning at each step is shown on top right.

\paragraph{Trust Score Dynamics (Figure~\ref{fig:trust_screenshot}).}
During episodes, agents are asked to output constantly explicit trust scores $T_{i \rightarrow j} \in [0,1]$ for all other players, representing their belief that player $j$ is a crewmate (higher values indicate greater trust). The visualization plots these scores over time, with each line tracking one agent's trust toward a specific player across successive voting rounds. 

\paragraph{Voting Interface (Figure~\ref{fig:voting_screenshot}).}
Voting phases are triggered on a fixed schedule. During voting, each alive agent generates a natural language statement defending themselves, and vote for the most suspicious player to eject. The visualization displays each agent's statement, their vote target, and the aggregated vote tally. The player receiving the most votes is ejected from the game (ties result in no ejection). This interface enables analysis of persuasion strategies, accusation patterns, and whether agents' stated reasoning aligns with their voting behavior.

\section{Prompting Strategy}
\label{app:prompts}

\paragraph{Movement System Prompt.}
The system prompt for crewmate movement is structured to provide comprehensive guidance for navigation and survival. It begins with a role description explaining the crewmate's objectives and victory conditions. The coordinate system is explicitly defined with the origin at the top-left corner, x-axis increasing rightward, and y-axis increasing downward. Action mechanics detail which tiles are blocked and the requirements for interactions such as opening doors or completing tasks. Safety rules instruct agents to assess threats based on proximity to other players, treating nearby agents as potential impostors. Anti-oscillation rules discourage repetitive back-and-forth movement patterns, while anti-stall rules penalize excessive NOOP actions when progress is possible. Finally, the output format specification requires agents to produce JSON responses containing both a reasoning trace (``thought'') and the selected action.

Impostor prompts build upon the crewmate template with additional role-specific content. Stealth mechanics guide agents to assess witness risk before attempting kills to avoid detection. Blending strategies encourage fake task interactions when other players are nearby to avoid suspicion. Kill cooldown information helps impostors plan their actions around the cooldown timer. Teammate identification prevents impostors from accidentally voting against fellow impostors during discussion phases.

\paragraph{Voting System Prompt.}
During voting phases, agents receive prompts tailored for social deduction. Trust scoring guidelines define the scale from 0.0 (definitely impostor) to 1.0 (definitely crewmate), with 0.5 representing uncertainty. Behavioral indicators highlight suspicious patterns such as proximity to dead bodies, erratic movement, or false accusations. The output format requires agents to provide their reasoning, a vote target (or skip), and trust scores for all alive players.

\subsection{Prompt Templates}
\label{app:prompt_templates}

We provide condensed prompt templates below, where \texttt{\{VAR\}} denotes dynamically populated fields. Full prompts are available in our released codebase.

\paragraph{Movement System Prompt (Crewmate).}
The movement system prompt follows this structure:

\begin{lstlisting}
You are a *Crewmate* in a {WIDTH}x{HEIGHT} grid.
Game Setup: {N_AGENTS} players, {N_IMPOSTORS} Impostor(s).

Primary Goals:
- SURVIVE and IDENTIFY Imposters.
- Victory: Vote out ALL Imposters OR complete YOUR tasks.

Global Actions: 0 MOVE_FORWARD, 1 MOVE_BACKWARD, 2 STRAFE_RIGHT,
  3 STRAFE_LEFT, 4-6 TURN_*, 7 NOOP, 8 DO_TASK, 9-10 DOOR, 
  11 REPORT_DEADBODY, 12 CALL_DISCUSSION

Legend: "#"=wall, "."=floor, "T"=task, "O/D"=door, 
        "C"=corpse, "B"=button, "P"=player

Coordinate System: origin=top_left, x_right, y_down
Position format: "Rxy(col,row)", e.g., "R00(5,3)"

Mechanics:
- Pick exactly ONE action from ALLOWED_ACTIONS.
- BLOCKED tiles: {"#","T","D","C","B","P"}
- Safety: Treat nearby players as potential threats.
- Anti-oscillation: Avoid undoing last move.

Output: {"thought":"<reasoning>","action":<int>}
\end{lstlisting}

\paragraph{Movement System Prompt (Impostor).}
Impostor prompts extend the crewmate template:

\begin{lstlisting}
You are an *Imposter* in a {WIDTH}x{HEIGHT} grid.

Primary Goals:
- ELIMINATE Crewmates and BLEND IN to avoid detection.
- Victory: Imposter_Count >= Crewmate_Count

Additional: 13 KILL (front_tile="P", cooldown=0)

Stealth Mechanics:
- WITNESS RISK: player within Manhattan dist<=2 can see.
- Kill SAFE only if zero witnesses; else reposition.
- Blending: DO_TASK on "T" tiles to fake progress.

Teammates: [{TEAMMATE_IDS}] - avoid voting against them.
Kill Cooldown: {COOLDOWN} steps remaining.

Output: {"thought":"<reasoning>","action":<int>}
\end{lstlisting}

\paragraph{User Prompt (Movement).}
Each step, the agent receives current state:

\begin{lstlisting}
SELF INFORMATION:
- NAME: {AGENT_ID}
- Position: {POSITION} (e.g., "R01(12,8)")
- Direction: {DIR} (100=R, 101=D, 102=L, 103=U)
- TASKS_TO_COMPLETE: [{name, pos, progress}, ...]
- ALLOWED_ACTIONS: [0,4,5,6,7,8]
- BEST_ACTION_SUGGESTION: [8] (from A* planner)
- TASK_NAV_ASTAR_DIST: {"short_task_1": 3, ...}
- NEAREST_PLAYER_DIST: 5

SELF HISTORY (last 5 steps):
  step=42: R01(12,9)->act=0->R01(12,8) ...
STALL_ALERT: false

OBSERVATION: {"step":45, "view":{"ascii_patch":{...}}}
\end{lstlisting}

\paragraph{Voting System Prompt.}
During voting phases:

\begin{lstlisting}
You are a *{ROLE}* named {AGENT_ID}.
Goal: Identify Imposters (Crew) / Frame innocents (Imp)

All Players: [Player_0, Player_1, ...]
Allowed Votes: [Player_0, ..., skip]

Trust Scoring:
- 0.0-0.3: Near body, erratic, false accusations
- 0.4-0.6: No info or mixed signals  
- 0.7-1.0: Verified tasks or cleared by logic

Voting: Vote LOWEST trust; if uncertain, "skip".

Output: {"thought":"...", "vote":"Player_X",
         "trust_scores":{"Player_0":0.8,...}}
\end{lstlisting}

\paragraph{Example Outputs.}
Movement response:
\begin{lstlisting}
{"thought":"Front=short_task_1; DO_TASK. No players 
  nearby (dist=5), safe. action=8","action":8}
\end{lstlisting}

Voting response:
\begin{lstlisting}
{"thought":"Player_2 near body, didn't report. 
  Player_0 did tasks. Player_1 false accusation.",
 "vote":"Player_2",
 "trust_scores":{"Player_0":0.9,"Player_1":0.4,
                 "Player_2":0.1}}
\end{lstlisting}

\subsection{Trust Assessment Protocol}
\label{app:trust_protocol}

Trust scores are collected at each voting phase throughout the game. Each agent outputs trust scores $T_{i \rightarrow j}(t) \in [0,1]$ for all other alive players, representing their assessment of each player's likelihood of being a crewmate. These scores are logged with timestamps to enable temporal analysis of how trust evolves over the course of an episode. The Brier score (Eq.~\ref{eq:trust_brier_score}) and volatility (Eq.~\ref{eq:trust_volatility}) metrics are computed post-hoc from these logged trust trajectories.

\section{Supplementary Details on Evaluation Metrics}
\label{app:metric_definitions}

This appendix provides implementation details for the evaluation metrics formally defined in Section~\ref{sec:em2bench}. We describe how each metric is computed in practice and discuss important considerations for interpreting the results.

\subsection{Planning Performance}
\label{app:planning_performance}

Planning Efficiency $\mathrm{PE}_{i,k}$ measures how close an agent's actual trajectory is to the optimal path for completing task $k$. Let $L^{\mathrm{opt}}_{i,k}$ denote the optimal path length and $L^{\mathrm{act}}_{i,k}$ denote the actual path length. The planning efficiency is then defined as:
\begin{equation}
    \mathrm{PE}_{i,k} = \frac{L^{\mathrm{opt}}_{i,k}}{L^{\mathrm{act}}_{i,k}} \in [0, 1].
\end{equation}

The optimal path length $L^{\mathrm{opt}}_{i,k}$ is computed as the sum of two components: the navigation cost and the interaction cost. The navigation cost corresponds to the A* shortest-path distance from the agent's starting position to the task location, accounting for walls, doors, and other obstacles in the gridworld. The interaction cost represents the number of toggle actions required to complete the task, which varies by task type: 3 toggles for common tasks, 8 for short tasks, and 13 for long tasks.

The actual path length $L^{\mathrm{act}}_{i,k}$ is defined as the total number of environment steps taken by agent $i$ from episode start until task $k$ is completed. This count includes all actions performed by the agent, encompassing movement commands, rotation actions, failed interactions, and NOOP commands. If task $k$ is not completed by the end of the episode, we set $\mathrm{PE}_{i,k} = 0$ to ensure that agents are not rewarded for partial progress without completion.

\subsection{Detection Accuracy Baseline}
\label{app:detection_baseline}

We report a static random baseline of $33\%$ for detection accuracy. In our default configuration with 7 players consisting of 5 crewmates and 2 impostors, a crewmate casting a non-skip vote chooses among 6 other players. Under uniform random voting, the probability of correctly selecting an impostor is:
\begin{equation}
    P(\text{correct}) = \frac{|\mathcal{I}|}{N - 1} = \frac{2}{6} \approx 33\%.
\end{equation}

It is important to note that as crewmates are eliminated during the game, the true chance baseline increases dynamically. For instance, the baseline rises to $2/5 = 40\%$ after one crewmate elimination and to $2/4 = 50\%$ after two eliminations. Nevertheless, we adopt the static $33\%$ baseline throughout our analysis for two reasons. First, the majority of voting phases in our experiments occur during the early game when few eliminations have taken place, making the initial ratio a reasonable approximation. Second, using the initial baseline provides a more conservative comparison that strengthens our negative results, as models fail to exceed even the most favorable random baseline.

Detection accuracy consistently hovers near this baseline across all experimental conditions and model scales, indicating that the gap to chance---whether computed statically or dynamically---remains negligible.

\section{Detailed Failure Analysis}
\label{app:failure_analysis}

\begin{figure}[h]
    \centering
    \includegraphics[width=0.95\linewidth]{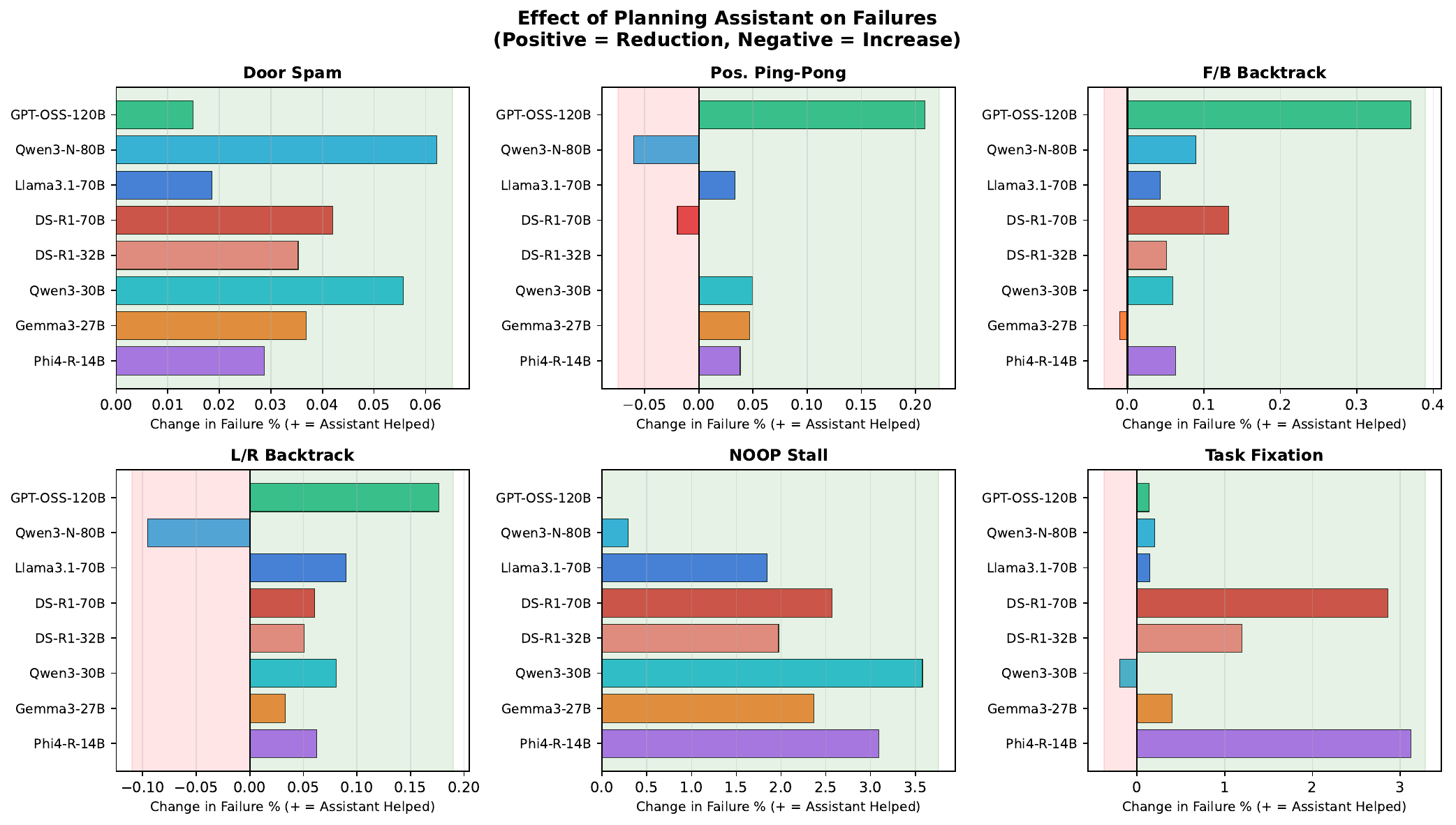}
    \caption{\textbf{Effect of Planning Assistant on Failure Modes.} Per-model change in failure percentage when the planning assistant is enabled. Positive values (green) indicate failures \emph{reduced} by the assistant; negative values (red) indicate failures that \emph{increase}. The assistant dramatically reduces passive failures (NOOP deadlock, task fixation) across most models, while active navigation failures show model-dependent effects.}
    \label{fig:failure_assist_effect}
\end{figure}

This appendix provides detailed quantitative results supporting the failure analysis presented in Section~\ref{sec:experiments}. Table~\ref{tab:failure_detailed} presents the complete breakdown of failure modes by model under both low (no assistance) and high (with planning oracle) conditions. Figure~\ref{fig:failure_assist_effect} visualizes the per-model effect of the planning assistant on each failure type, revealing which failure modes are most effectively mitigated by external guidance.

\begin{table}[h]
    \centering
    \caption{Detailed Failure Mode Analysis by Model. Values show fraction of episode steps (\%) for each failure pattern. Color gradient: \colorbox[RGB]{180,220,180}{green} = low (good), \colorbox[RGB]{255,100,100}{red} = high (bad).}
    \label{tab:failure_detailed}
    \scriptsize
    \setlength{\tabcolsep}{3pt}
    \begin{tabular}{lccccccccc}
        \toprule
        \textbf{Model}  & \textbf{C} & \rotatebox{70}{\textbf{Door-Sp}} & \rotatebox{70}{\textbf{Pos.P.P.}} & \rotatebox{70}{\textbf{F/B-BT}} & \rotatebox{70}{\textbf{L/R-BT}} & \rotatebox{70}{\textbf{NOOP}}  & \rotatebox{70}{\textbf{Task-Fix}} & \rotatebox{70}{\textbf{Stuck}}   \\
        \midrule
        Phi4-R 14B  & Low & \cellcolor[RGB]{147,220,75}0.1 & \cellcolor[RGB]{97,220,111}0.1 & \cellcolor[RGB]{102,220,107}0.1 & \cellcolor[RGB]{111,220,101}0.1 & \cellcolor[RGB]{255,67,27}3.2 & \cellcolor[RGB]{255,0,0}3.2 & \cellcolor[RGB]{255,15,6}0.06 \\
                    & High & \cellcolor[RGB]{75,220,126}0.0 & \cellcolor[RGB]{27,220,160}0.0 & \cellcolor[RGB]{26,220,160}0.0 & \cellcolor[RGB]{29,220,159}0.0 & \cellcolor[RGB]{12,220,171}0.1 & \cellcolor[RGB]{17,220,167}0.1 & \cellcolor[RGB]{18,220,166}0.00 \\
        Gemma3 27B  & Low & \cellcolor[RGB]{184,220,49}0.1 & \cellcolor[RGB]{121,220,94}0.1 & \cellcolor[RGB]{128,220,89}0.1 & \cellcolor[RGB]{74,220,127}0.1 & \cellcolor[RGB]{255,162,66}2.4 & \cellcolor[RGB]{79,220,123}0.5 & \cellcolor[RGB]{254,220,0}0.03 \\
                    & High & \cellcolor[RGB]{91,220,115}0.0 & \cellcolor[RGB]{36,220,154}0.0 & \cellcolor[RGB]{140,220,80}0.1 & \cellcolor[RGB]{31,220,157}0.0 & 0.0 & \cellcolor[RGB]{16,220,168}0.1 & \cellcolor[RGB]{16,220,168}0.00 \\
        Qwen3 30B   & Low & \cellcolor[RGB]{180,220,52}0.1 & \cellcolor[RGB]{118,220,96}0.1 & \cellcolor[RGB]{125,220,91}0.1 & \cellcolor[RGB]{136,220,83}0.1 & \cellcolor[RGB]{255,0,0}3.7 & \cellcolor[RGB]{60,220,137}0.4 & \cellcolor[RGB]{255,120,49}0.04 \\
                    & High & \cellcolor[RGB]{40,220,151}0.0 & \cellcolor[RGB]{29,220,159}0.0 & \cellcolor[RGB]{54,220,141}0.0 & \cellcolor[RGB]{30,220,158}0.0 & \cellcolor[RGB]{23,220,163}0.2 & \cellcolor[RGB]{91,220,115}0.6 & \cellcolor[RGB]{55,220,140}0.01 \\
        DS-R1 32B   & Low & \cellcolor[RGB]{152,220,72}0.1 & 0.0 & \cellcolor[RGB]{106,220,104}0.1 & \cellcolor[RGB]{115,220,98}0.1 & \cellcolor[RGB]{255,45,18}3.4 & \cellcolor[RGB]{255,217,88}1.6 & \cellcolor[RGB]{255,87,35}0.05 \\
                    & High & \cellcolor[RGB]{64,220,134}0.0 & 0.0 & \cellcolor[RGB]{44,220,148}0.0 & \cellcolor[RGB]{48,220,145}0.0 & \cellcolor[RGB]{189,220,46}1.4 & \cellcolor[RGB]{69,220,130}0.4 & \cellcolor[RGB]{154,220,70}0.02 \\
        DS-R1 70B   & Low & \cellcolor[RGB]{158,220,68}0.1 & 0.0 & \cellcolor[RGB]{210,220,31}0.2 & \cellcolor[RGB]{119,220,95}0.1 & \cellcolor[RGB]{255,28,11}3.5 & \cellcolor[RGB]{255,29,12}3.0 & \cellcolor[RGB]{255,0,0}0.06 \\
                    & High & \cellcolor[RGB]{53,220,142}0.0 & \cellcolor[RGB]{35,220,154}0.0 & \cellcolor[RGB]{50,220,144}0.0 & \cellcolor[RGB]{40,220,151}0.0 & \cellcolor[RGB]{127,220,90}0.9 & \cellcolor[RGB]{24,220,162}0.2 & \cellcolor[RGB]{97,220,111}0.01 \\
        Llama3.1 70B & Low & \cellcolor[RGB]{123,220,92}0.0 & \cellcolor[RGB]{117,220,96}0.1 & \cellcolor[RGB]{105,220,105}0.1 & \cellcolor[RGB]{228,220,18}0.2 & \cellcolor[RGB]{250,220,2}1.8 & \cellcolor[RGB]{55,220,140}0.4 & \cellcolor[RGB]{203,220,36}0.02 \\
                    & High & \cellcolor[RGB]{77,220,125}0.0 & \cellcolor[RGB]{56,220,139}0.0 & \cellcolor[RGB]{53,220,142}0.0 & \cellcolor[RGB]{110,220,101}0.1 & 0.0 & \cellcolor[RGB]{32,220,157}0.2 & \cellcolor[RGB]{22,220,164}0.00 \\
        Qwen3 80B   & Low & \cellcolor[RGB]{255,0,0}0.2 & \cellcolor[RGB]{255,121,49}0.2 & \cellcolor[RGB]{255,133,54}0.3 & \cellcolor[RGB]{255,107,44}0.3 & \cellcolor[RGB]{39,220,151}0.3 & \cellcolor[RGB]{218,220,25}1.4 & \cellcolor[RGB]{159,220,67}0.02 \\
                    & High & \cellcolor[RGB]{255,134,54}0.1 & \cellcolor[RGB]{255,27,11}0.3 & \cellcolor[RGB]{246,220,5}0.2 & \cellcolor[RGB]{255,0,0}0.4 & 0.0 & \cellcolor[RGB]{185,220,48}1.2 & \cellcolor[RGB]{122,220,93}0.01 \\
        GPT-OSS 120B & Low & \cellcolor[RGB]{255,173,71}0.1 & \cellcolor[RGB]{255,0,0}0.3 & \cellcolor[RGB]{255,0,0}0.4 & \cellcolor[RGB]{255,181,74}0.2 & 0.0 & \cellcolor[RGB]{63,220,134}0.4 & \cellcolor[RGB]{78,220,124}0.01 \\
                    & High & \cellcolor[RGB]{255,206,84}0.1 & \cellcolor[RGB]{131,220,86}0.1 & \cellcolor[RGB]{61,220,136}0.1 & \cellcolor[RGB]{68,220,131}0.1 & 0.0 & \cellcolor[RGB]{41,220,150}0.3 & \cellcolor[RGB]{29,220,159}0.00 \\
        \bottomrule
    \end{tabular}
    \vspace{1mm}
    \parbox{\linewidth}{\scriptsize
        \textbf{Legend:} C = Condition (Low = no assistance, High = with oracle). Columns: Door Spam, Position Ping-Pong, Forward/Back Backtrack, Left/Right Backtrack, NOOP Deadlock, Task Fixation, Stuckness Index.}
\end{table}

\section{Social Reasoning Failure Analysis}
\label{app:social_failure_analysis}

This appendix documents the methodology and detailed findings for the social reasoning failure analysis presented in the main text. We analyze 64,184 crewmate votes and 28,158 impostor votes from voting logs across all league matchups. We extracted the ``thought'' field (natural language reasoning) and ``trust\_scores'' (numerical trust assignments) from each voting log entry. To identify failure modes, we developed pattern-based classifiers through an empirical, data-driven process.

\paragraph{Step 1: Keyword Frequency Analysis.}
We first computed the frequency of candidate keywords across all 92,342 reasoning traces (64,184 crewmate + 28,158 impostor). Table~\ref{tab:keyword_freq} shows the most frequent patterns, revealing that superficial behavioral cues (``erratic,'' ``stationary,'' ``movement pattern'') dominate crewmate reasoning.

\begin{table}[h]
\centering
\small
\caption{\textbf{Keyword frequency in crewmate reasoning traces.} Top patterns extracted from 92,342 voting log entries.}
\label{tab:keyword_freq}
\begin{tabular}{lrr}
\toprule
\textbf{Keyword/Pattern} & \textbf{Count} & \textbf{\% of Votes} \\
\midrule
erratic & 13,761 & 67.8\% \\
consistent & 9,808 & 48.4\% \\
no suspicious & 5,420 & 26.7\% \\
not doing tasks & 5,159 & 25.4\% \\
body & 4,715 & 23.2\% \\
near body & 4,189 & 20.7\% \\
movement pattern & 3,712 & 18.3\% \\
loop & 3,711 & 18.3\% \\
stationary & 2,969 & 14.6\% \\
no evidence & 1,870 & 9.2\% \\
equal/tied trust & 1,631 & 8.0\% \\
kill & 338 & 1.7\% \\
vent & 164 & 0.8\% \\
arbitrary & 95 & 0.5\% \\
\bottomrule
\end{tabular}
\end{table}

\paragraph{Step 2: Pattern-Accuracy Correlation.}
We computed the detection accuracy (probability of correctly voting for an impostor) conditioned on each keyword's presence or absence. Table~\ref{tab:pattern_accuracy} reveals which patterns actually help identify impostors versus which are misleading heuristics.

\begin{table}[h]
\centering
\small
\caption{\textbf{Detection accuracy by reasoning pattern.} Accuracy when pattern is present vs.\ absent. Patterns that \emph{hurt} accuracy (negative delta) indicate weak or misleading heuristics.}
\label{tab:pattern_accuracy}
\begin{tabular}{lccc}
\toprule
\textbf{Pattern} & \textbf{Acc.\ (Present)} & \textbf{Acc.\ (Absent)} & \textbf{Delta} \\
\midrule
\multicolumn{4}{l}{\textit{Strong Evidence (helps)}} \\
\quad kill & 53.3\% & 11.9\% & \textcolor{green!60!black}{+41.4\%} \\
\quad saw/witnessed & 19.4\% & 12.4\% & \textcolor{green!60!black}{+7.0\%} \\
\quad vent & 17.1\% & 12.6\% & \textcolor{green!60!black}{+4.5\%} \\
\midrule
\multicolumn{4}{l}{\textit{Weak Heuristics (hurts or no help)}} \\
\quad stationary & 8.5\% & 13.3\% & \textcolor{red}{-4.8\%} \\
\quad loop/oscillating & 9.3\% & 13.7\% & \textcolor{red}{-4.4\%} \\
\quad near body & 10.0\% & 13.3\% & \textcolor{red}{-3.3\%} \\
\quad erratic & 12.4\% & 12.9\% & -0.5\% \\
\quad movement pattern & 11.8\% & 12.8\% & -1.0\% \\
\midrule
\multicolumn{4}{l}{\textit{Evidence Scarcity (hurts)}} \\
\quad arbitrary & 0.0\% & 12.6\% & \textcolor{red}{-12.6\%} \\
\quad no evidence & 9.1\% & 12.9\% & \textcolor{red}{-3.8\%} \\
\quad equal/tied trust & 9.9\% & 12.8\% & \textcolor{red}{-2.9\%} \\
\bottomrule
\end{tabular}
\end{table}

\paragraph{Step 3: Classification Rules.}
Based on the empirical analysis, we defined three failure mode categories using regular expression patterns. \emph{Evidence Scarcity} captures patterns indicating lack of evidence and arbitrary voting (e.g., ``arbitrar,'' ``equal.*trust,'' ``no.*evidence''). \emph{Weak Heuristics} captures superficial behavioral cues that hurt or provide no accuracy benefit (e.g., ``stationary,'' ``loop,'' ``erratic,'' ``near.*body''). \emph{Strong Evidence} captures concrete observations that actually help identify impostors (e.g., ``kill,'' ``vent,'' ``saw''); votes with strong evidence are excluded from failure modes. For impostor votes, we classified \emph{Misdirection} as cases where the impostor voted for a crewmate while assigning trust $\geq 0.7$ to their fellow impostor, indicating intentional teammate protection. Table~\ref{tab:crewmate_detailed} shows the full breakdown of crewmate votes by failure category.

\begin{table}[h]
\centering
\small
\caption{\textbf{Crewmate vote classification.} Analysis of 64,184 non-skip votes.}
\label{tab:crewmate_detailed}
\begin{tabular}{lrrr}
\toprule
\textbf{Category} & \textbf{Count} & \textbf{\% of Votes} & \textbf{Accuracy} \\
\midrule
Evidence Scarcity & 10,077 & 15.7\% & 32.8\% \\
Weak Heuristics & 40,308 & 62.8\% & 43.9\% \\
Strong Evidence & 8,024 & 12.5\% & 48.2\% \\
Other & 5,775 & 9.0\% & 31.5\% \\
\midrule
\textbf{Overall} & 64,184 & 100\% & 40.2\% \\
\bottomrule
\end{tabular}
\end{table}

\paragraph{Impostor Vote Classification.}
Table~\ref{tab:impostor_detailed} shows impostor voting behavior. Notably, impostors \emph{never} vote for their teammates, and the majority successfully misdirect while protecting their partner.

\begin{table}[h]
\centering
\small
\caption{\textbf{Impostor vote classification.} Analysis of 28,158 non-skip votes.}
\label{tab:impostor_detailed}
\begin{tabular}{lrr}
\toprule
\textbf{Category} & \textbf{Count} & \textbf{\% of Votes} \\
\midrule
Misdirection (protecting teammate) & 23,737 & 84.3\% \\
Framed crewmate (no protection) & 4,421 & 15.7\% \\
Voted for teammate & 0 & 0.0\% \\
\bottomrule
\end{tabular}
\end{table}

\paragraph{Over-trust Analysis.}
We additionally analyzed how often crewmates assign high trust to impostors. Of 64,184 crewmate votes, 46.7\% (30,005 votes) assigned trust $T \geq 0.7$ to at least one impostor. These over-trusting votes achieved only 33.4\% detection accuracy, compared to 43.1\% overall. The mean trust score assigned to impostors was 0.532, with 32.8\% of impostor trust scores falling in the high-trust range ($T \geq 0.7$). This indicates that impostors successfully blend in, and crewmates frequently fail to identify suspicious behavior.

\paragraph{Key Findings.}
\begin{enumerate}
\item \textbf{Weak heuristics dominate reasoning}: 62.8\% of crewmate votes rely on superficial cues like ``erratic movement'' that provide limited accuracy benefit.
\item \textbf{Evidence scarcity leads to below-average performance}: When crewmates explicitly acknowledge lack of evidence (15.7\% of votes), their accuracy drops to 32.8\%---below the 43.1\% average.
\item \textbf{Over-trust is pervasive}: 46.7\% of votes assign high trust ($T \geq 0.7$) to impostors, dropping accuracy to 33.4\%.
\item \textbf{Strong evidence is rare but effective}: Only 12.8\% of votes cite concrete evidence (witnessing kills, vents), but these achieve 54.6\% accuracy---significantly above baseline.
\item \textbf{Impostors successfully coordinate}: 84.3\% of impostor votes protect teammates while targeting crewmates, and 0\% vote for their partner.
\end{enumerate}

\section{Complexity Analysis of Task and Planning Performance}
\label{app:complexity_analysis_tp_pp}

\begin{figure}[h]
    \centering
    \begin{subfigure}[t]{0.80\linewidth}
        \centering
        \includegraphics[width=.9\linewidth]{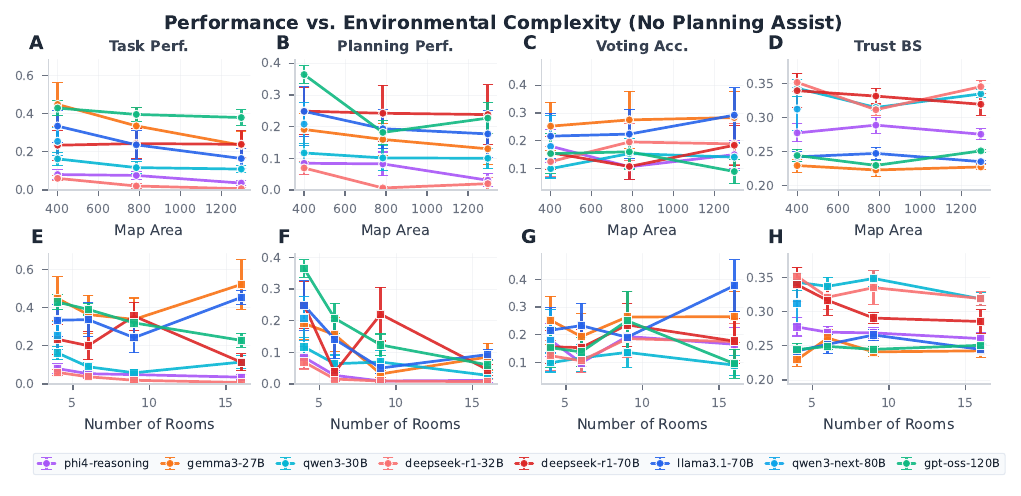}
        \caption{\textbf{No Planning Assistance.} Performance degrades sharply with increasing complexity.}
        \label{fig:complexity_analysis_tp_pp_noassist}
    \end{subfigure}
    \vspace{0.5em}
    \begin{subfigure}[t]{0.80\linewidth}
        \centering
        \includegraphics[width=.9\linewidth]{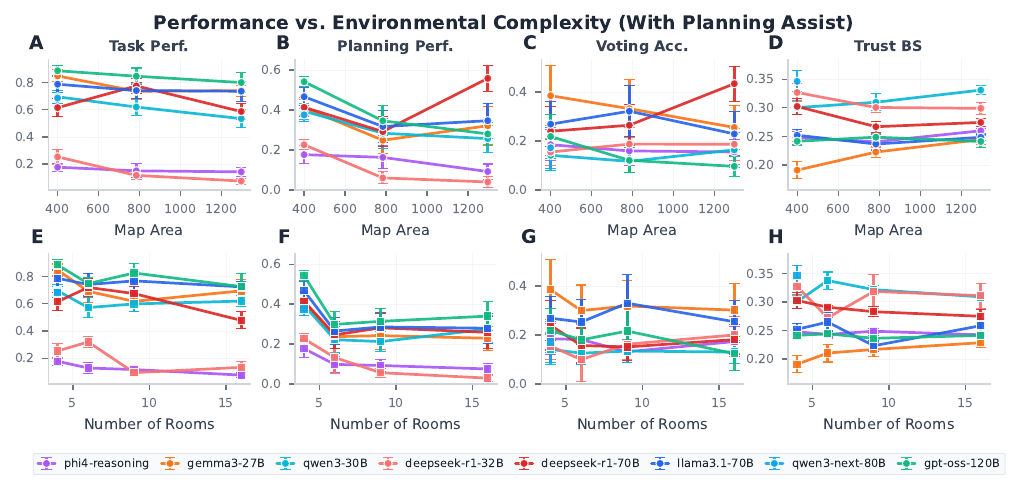}
        \caption{\textbf{With Planning Assistance.} Performance improves substantially; degradation is attenuated.}
        \label{fig:complexity_analysis_tp_pp_assist}
    \end{subfigure}
    \caption{\textbf{Performance vs.\ environmental complexity.} Top row: performance vs.\ map area (fixed $2{\times}2$ layout). Bottom row: performance vs.\ number of rooms (fixed $10{\times}10$ room size). Columns show Task Performance (TP), Planning Performance (PP), Voting Accuracy, and Trust Brier Score (BS). Error bars indicate standard error across episodes.}
    \label{fig:complexity_analysis_tp_pp_compare}
\end{figure}

We analyze how model performance scales with environmental complexity along two orthogonal dimensions: \textbf{room size} (map area with fixed $2{\times}2$ layout) and \textbf{room layout} (number of rooms with fixed $10{\times}10$ room size). Figure~\ref{fig:complexity_analysis_tp_pp_compare} presents this analysis for both conditions.

\paragraph{Task Performance (TP) and Planning Performance (PP).}
Without planning assistance (Figure~\ref{fig:complexity_analysis_tp_pp_noassist}), both TP and PP degrade substantially as map area increases, with all models converging toward near-zero performance on larger maps. GPT-OSS-120B maintains the highest performance across scales, but even it drops below 20\% task completion on the largest configurations. The room layout dimension shows similar trends: increasing the number of rooms (while holding room size constant) degrades performance, though the effect is less pronounced than map area scaling.

With planning assistance (Figure~\ref{fig:complexity_analysis_tp_pp_assist}), performance improves dramatically across all models and complexity levels. TP increases from $<$20\% to 60--80\% on smaller maps, and the degradation with increasing complexity is attenuated. Planning Performance shows the most striking improvement: PP values approach 0.8--0.9 with assistance versus 0.1--0.3 without, confirming that the planning oracle successfully addresses the navigation bottleneck identified in Section~\ref{sec:experiments}.

\paragraph{Voting Accuracy and Trust Calibration.}
Interestingly, voting accuracy and trust Brier score (BS) show minimal sensitivity to environmental complexity in both conditions. Detection accuracy hovers near the random baseline (33\%) regardless of map size or room count, reinforcing our finding that social reasoning limitations are not attributable to navigation difficulty. Even when planning assistance removes the spatial reasoning burden, models fail to improve their impostor detection capabilities.

\section{Environment Configuration}
\label{app:env_config}
\paragraph{Default Environment Parameters.}
Unless otherwise specified, all experiments use a default room pattern of $10 \times 10$ grid size with $2 \times 1$ layout, which consists of 4 rooms arranged in a $2 \times 2$ layout with each room spanning $10 \times 10$ tiles. The environment hosts 7 players in total, including 2 impostors and 5 crewmates. Each agent observes the environment through a limited field of view with a radius of 4 tiles centered on their position. Episodes run for a maximum of 500 macro-steps, where each macro-step consists of 5 environment steps (totaling 2,500 environment steps). In addition to event-triggered voting (dead body reports or emergency button presses), the environment schedules voting sessions every 200 environment steps (i.e., every 40 macro-steps) to ensure regular social reasoning opportunities. At the start of each episode, agents are placed at random positions within the gridworld.

\paragraph{Task Configuration.}
The environment features three types of tasks with varying completion requirements. Common tasks require 3 toggle actions to complete and yield a reward of $+1.0$ upon completion. Short tasks require 8 toggle actions and provide a reward of $+2.0$. Long tasks are the most demanding, requiring 13 toggle actions with a reward of $+3.0$. By default, each crewmate is assigned one task of each type at the beginning of an episode.

\paragraph{Complexity Sweep Parameters.}
For the systematic complexity analysis described in Section~\ref{sec:experiments}, we vary two primary dimensions. The map area ranges from $10 \times 10$ to $14 \times 14$ and $18 \times 18$ grid sizes. The room count varies across $2 \times 2$, $2 \times 3$, $3 \times 3$, and $4 \times 4$ layouts, allowing us to assess how spatial complexity affects agent performance.

\section{LLM Agent Configuration}
\label{app:llm_config}

\paragraph{Models Evaluated.}
Our experiments encompass both standard and reasoning-specialized models. The standard models include GPT-OSS-120B, Llama3.1-70B, Qwen3-next-80B, Qwen3-30B, and Gemma3-27B. We also evaluate reasoning models specifically designed for complex inference tasks: DeepSeek-R1-70B and Phi4-Reasoning-14B. All models are publicly available via the Ollama model library for full reproducibility, as shown in Table~\ref{tab:model_urls}.

\begin{table}[h]
\centering
\caption{\textbf{Model Availability.} All evaluated models are publicly accessible via Ollama.}
\label{tab:model_urls}
\begin{tabular}{lll}
\toprule
\textbf{Model} & \textbf{Ollama Tag} & \textbf{URL} \\
\midrule
GPT-OSS-120B & \texttt{gpt-oss:120b} & \url{https://ollama.com/library/gpt-oss} \\
Qwen3-next-80B & \texttt{qwen3-next:80b} & \url{https://ollama.com/library/qwen3-next} \\
Llama3.1-70B & \texttt{llama3.1:70b} & \url{https://ollama.com/library/llama3.1} \\
DeepSeek-R1-70B & \texttt{deepseek-r1:70b} & \url{https://ollama.com/library/deepseek-r1} \\
Qwen3-30B & \texttt{qwen3:30b} & \url{https://ollama.com/library/qwen3} \\
Gemma3-27B & \texttt{gemma3:27b} & \url{https://ollama.com/library/gemma3} \\
Phi4-Reasoning-14B & \texttt{phi4:latest} & \url{https://ollama.com/library/phi4} \\
\bottomrule
\end{tabular}
\end{table}

\paragraph{Inference Configuration.}
All LLM inference is performed using vLLM~\cite{kwon2023vllm} for efficient batched generation on NVIDIA H100 80GB GPUs. We use a temperature of 0 (greedy decoding) to ensure reproducibility across runs. The maximum output token limit is set to 2048 tokens, and each LLM call has a timeout of 240 seconds. Responses are constrained to structured JSON format with Pydantic schema validation to ensure parseable outputs. For GPT-OSS models, we additionally configure \texttt{reasoning\_effort="low"} to control deliberation depth, while DeepSeek and Qwen models use \texttt{reasoning\_effort="none"}.

\paragraph{Observation Format.}
Each agent receives observations in JSON format containing several key components. The primary visual input is an ASCII minimap patch consisting of a $9 \times 9$ local view centered on the agent, with visibility occlusion applied to simulate realistic perception. Position information encodes the agent's current grid coordinates using a room-relative format (e.g., ``R00(5,3)'' indicates row 0, column 0 of the room grid at coordinates $(5,3)$), along with facing direction encoded as integers 100--103 representing RIGHT, DOWN, LEFT, and UP respectively. The observation also includes a list of assigned tasks with their positions, current progress, and completion status, a dictionary mapping visible positions to player identities, and a dynamically computed list of valid actions for the current state.

\paragraph{Action Space.}
Agents can perform 14 discrete actions. Movement actions include MOVE\_FORWARD (0), MOVE\_BACKWARD (1), STRAFE\_RIGHT (2), and STRAFE\_LEFT (3). Rotation actions comprise TURN\_LEFT (4), TURN\_RIGHT (5), and TURN\_BACK (6). Interaction actions include NOOP (7), DO\_TASK (8), OPEN\_DOOR (9), CLOSE\_DOOR (10), REPORT\_DEADBODY (11), and CALL\_DISCUSSION (12). Impostors have access to an additional action, KILL (13), which is unavailable to crewmates.

\subsection{Planning Assistant}
\label{app:planning_assistant}

The planning assistant provides navigation guidance using A* pathfinding to help agents navigate the gridworld efficiently. For task navigation, the assistant computes optimal paths to each assigned task and provides the suggested next action along with the A* distance. Player navigation (used primarily by impostors) computes paths to each visible crewmate for hunting purposes. Button navigation computes the path to the emergency discussion button, while dead body navigation provides paths to report visible corpses.

Three difficulty modes control the level of assistance provided in the prompt. In easy mode, agents receive full navigation suggestions with explicit best action recommendations. Medium mode provides navigation suggestions along with predicted action outcomes for spatial verification, requiring agents to reason about the suggested paths. Hard mode offers no navigation assistance, requiring agents to plan independently using only their observations. All experiments labeled as ``planning assistant enabled'' use the easy mode.

\section{Reward Function Details}
\label{app:reward_function}

The reward function in SocialGrid is designed to incentivize role-appropriate behavior while maintaining game balance between crewmates and impostors. Rewards are structured across three phases: task execution, voting, and game outcomes. Table~\ref{tab:reward_function} summarizes all reward components.

\paragraph{Task Phase Rewards.}
Crewmates receive incremental rewards for task progress: $+0.2$ per toggle action and completion bonuses scaled by task difficulty ($+1.0$ for common, $+2.0$ for short, $+3.0$ for long tasks). This encourages steady task completion rather than abandoning partially-finished tasks. Impostors receive $+6.0$ for successful kills, while victims incur $-6.0$, creating strong incentives for both hunting and survival. A small step penalty ($-0.001$ for crewmates, $-0.005$ for impostors) discourages idle behavior and encourages efficient movement.

\paragraph{Voting Phase Rewards.}
Correct impostor identification yields $+3.0$ for crewmates, while incorrect votes against fellow crewmates incur $-2.0$, encouraging careful deliberation. Skip votes receive a small positive reward ($+0.05$) to make abstention viable when uncertain. Ejected impostors receive $-3.0$, while wrongly ejected crewmates incur $-2.0$. Impostors gain $+2.0$ when a crewmate is ejected, rewarding successful misdirection.

\paragraph{Game Outcome Rewards.}
Terminal rewards of $\pm 10$ for win/loss dominate the reward signal, ensuring agents prioritize team victory over individual metrics. The asymmetric structure (crewmates gain from completing tasks, impostors from eliminating crewmates) creates natural tension that mirrors the social deduction dynamics of the original game.

\begin{table}[h]
    \caption{Reward function summary for \name. Rewards are role-specific: ``Crew'' applies to crewmates, ``Imp'' to impostors. Dashes indicate rewards that do not apply to that role.}
    \label{tab:reward_function}
    \vskip 0.1in
    \begin{center}
        \begin{small}
            \begin{sc}
                \begin{tabular}{lcc}
                    \toprule
                    Event & Crew & Imp \\
                    \midrule
                    \multicolumn{3}{l}{\textit{Task Phase}} \\
                    Task toggle & +0.2 & -- \\
                    Common task done & +1.0 & -- \\
                    Short task done & +2.0 & -- \\
                    Long task done & +3.0 & -- \\
                    Kill (killer) & -- & +6.0 \\
                    Kill (victim) & $-6.0$ & -- \\
                    Step penalty & $-0.001$ & $-0.005$ \\
                    \midrule
                    \multicolumn{3}{l}{\textit{Voting Phase}} \\
                    Vote imp (correct) & +3.0 & -- \\
                    Vote crew (wrong) & $-2.0$ & -- \\
                    Skip vote & +0.05 & +0.05 \\
                    Ejected (imp) & -- & $-3.0$ \\
                    Ejected (crew) & $-2.0$ & -- \\
                    Crew ejected (imp bonus) & -- & +2.0 \\
                    \midrule
                    \multicolumn{3}{l}{\textit{Game Outcome}} \\
                    Crewmates win & +10 & $-10$ \\
                    Impostors win & $-10$ & +10 \\
                    \bottomrule
                \end{tabular}
            \end{sc}
        \end{small}
    \end{center}
    \vskip -0.1in
\end{table}

\section{Training and Evaluation Protocol}
\label{app:training_eval}

\subsection{Reinforcement Learning Configuration}
\label{app:rl_config}

For the RL experiments addressing research question Q5 in Section~\ref{sec:experiments}, we employ Proximal Policy Optimization (PPO)~\cite{schulman2017proximal}. Table~\ref{tab:ppo_hyperparams} summarizes the hyperparameters used for training.

\begin{table}[h]
    \caption{PPO Hyperparameters for RL Training}
    \label{tab:ppo_hyperparams}
    \vskip 0.1in
    \begin{center}
        \begin{small}
            \begin{sc}
                \begin{tabular}{lc}
                    \toprule
                    Hyperparameter & Value \\
                    \midrule
                    Learning Rate & $3 \times 10^{-7}$ \\
                    Discount Factor ($\gamma$) & 0.99 \\
                    GAE Lambda ($\lambda$) & 0.95 \\
                    Clipping Coefficient & 0.2 \\
                    Entropy Coefficient & 0.01 \\
                    Value Function Coefficient & 0.5 \\
                    Maximum Gradient Norm & 1.0 \\
                    Number of Parallel Environments & 1 \\
                    \midrule
                    PPO Epochs per Update & 4 \\
                    Batch Size (per PPO Update) & 16 \\
                    Mini-batch Size & 4 \\
                    Gradient Accumulation Steps & 2 \\
                    Total PPO Updates & 2{,}500 \\
                    \bottomrule
                \end{tabular}
            \end{sc}
        \end{small}
    \end{center}
    \vskip -0.1in
\end{table}

\paragraph{RL Environment Settings.}
For RL training, we use a simplified environment configuration to reduce computational overhead. The room size is set to $7 \times 7$ tiles, and room layouts are sampled from ${1{\times}1, 1{\times}2, 1{\times}3, 2{\times}2}$ at the beginning of each episode. The environment hosts 1 players consisting of 1 crewmate and 0 impostor, and victory conditions are restricted to task completion only and episodes can run for up to 200 steps.

\paragraph{Base Model for RL Fine-tuning.}
We fine-tune Qwen3-4B-Instruct-2507 using the reward function described in Appendix~\ref{app:reward_function}.
To enable parameter-efficient reinforcement learning, we apply Low-Rank Adaptation (LoRA) to the base model, while keeping all original model parameters frozen. 
LoRA adapters are configured with rank 8, scaling factor 16, and dropout 0.05.
Checkpoints are saved every 300 PPO updates, and we report evaluation results at the baseline (step 0), an intermediate checkpoint (step 1200), and the final trained model.

\subsection{Episode and Evaluation Protocol}
\label{app:eval_protocol}

\paragraph{Number of Episodes.}
The number of evaluation episodes varies by experiment type. For the main model comparison addressing Q1, we run 20 episodes per model per condition (with and without the planning assistant). The complexity analysis for Q3 uses 3 episodes per model per environment configuration across all combinations of map sizes and room layouts. For league matching experiments, we collect 486 scored episodes on the $10{\times}10$\_$2{\times}2$ pattern and 466 scored episodes on the $14{\times}14$\_$2{\times}2$ pattern, totaling 952 episodes across 30 unique model matchups per pattern.

\paragraph{Episode Execution.}
Each episode is executed sequentially with agent responses generated in real-time to simulate authentic multi-agent interactions. Episodes terminate under one of five conditions: crewmates win if all impostors are successfully voted out or if all surviving crewmates complete their assigned tasks; impostors win if all crewmates are eliminated, if impostor parity is reached (number of impostors equals or exceeds number of crewmates), or if the maximum step limit is reached without crewmate victory.

\subsection{Failure Detection Parameters}
\label{app:failure_params}

The automatic failure detection system described in Section~\ref{sec:experiments} uses specific thresholds to identify problematic agent behaviors. Door toggle failures are flagged when an agent performs 3 or more door open/close actions on the same door within 10 steps. Door spam is detected when 5 or more door interactions occur within 15 steps without any position change. Position ping-pong identifies agents visiting 3 or fewer distinct positions 4 or more times within a 20-step window.

Movement-related failures include move backtrack loops (forward-then-backward patterns occurring 3 or more times within 15 steps), strafe backtrack loops (left-then-right patterns with the same threshold), move oscillation (alternating single forward/backward steps 4 or more times within 10 steps), and strafe oscillation (alternating single left/right steps with the same threshold). Turn toggle failures occur when an agent performs 4 or more turn actions without any movement within 8 steps.

Stall-type failures capture inactivity: NOOP stall is triggered by 5 or more consecutive NOOP actions, while task fixation stall is flagged when an agent performs 10 or more consecutive DO\_TASK actions without successfully completing any task.

\subsection{Winning Score Calculation Parameters}
\label{app:winning_score_params}

The winning score heuristic detailed in Appendix~\ref{app:winning_score_calculation} uses several configurable weights. The task completion contribution weight $w_{\text{task}}$ is set to 1.0, the alive-advantage contribution weight $w_{\text{alive}}$ is 0.5, and the detection signal contribution weight $w_{\text{detect}}$ is 0.3. The sigmoid steepness parameter $\alpha$ controlling the probability mapping is set to 2.0.

\subsection{Reproducibility}
\label{app:reproducibility}

To ensure reproducibility of our results, we employ several standardization measures. All random seeds are fixed at experiment start to ensure deterministic environment initialization and agent placement. LLM temperature is set to 0 for greedy decoding, eliminating sampling variance in model outputs. Environment configurations are fully specified in JSON files that can be loaded to recreate exact experimental conditions. Complete prompt templates are provided in our codebase, and all evaluated model weights are publicly available through their respective providers. The benchmark code, evaluation suite, and experimental logs are publicly available at \href{https://github.com/ml-research/SocialGrid}{github.com/ml-research/SocialGrid}.

\section{Head-to-Head League Match Heatmap}
\label{app:league_heatmap}

To provide a more detailed view of model performance in direct competition, we present a head-to-head heatmap visualization showing win rates between all model pairs. This complements the aggregate league rankings in Table~\ref{tab:league_rankings_full} by revealing asymmetric matchup dynamics---some model pairs exhibit strong dominance relationships that are obscured in aggregate statistics.

\begin{figure}[h]
    \centering
    \begin{minipage}{0.48\linewidth}
        \centering
        \includegraphics[width=\linewidth]{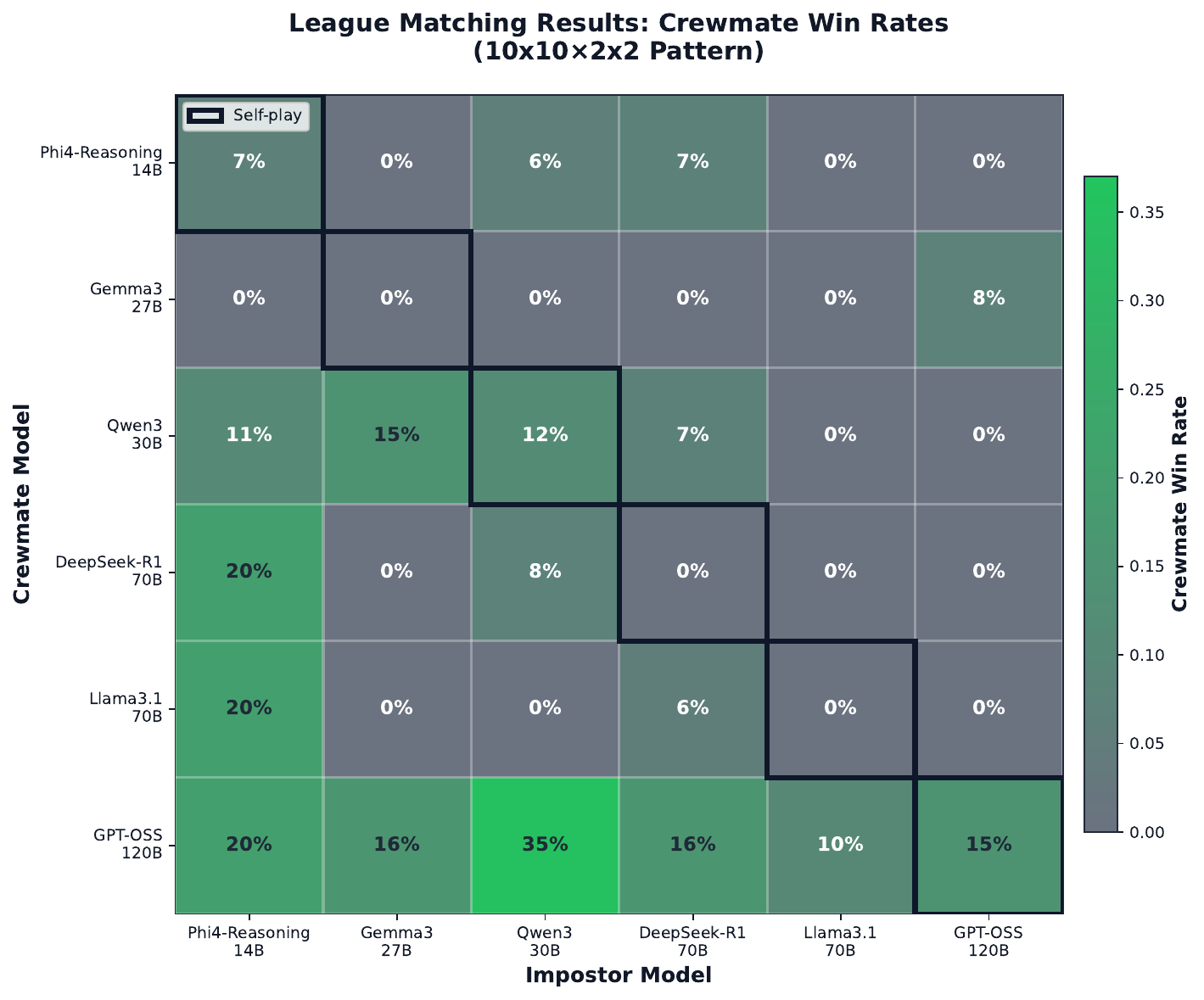}
    \end{minipage}
    \hfill
    \begin{minipage}{0.48\linewidth}
        \centering
        \includegraphics[width=\linewidth]{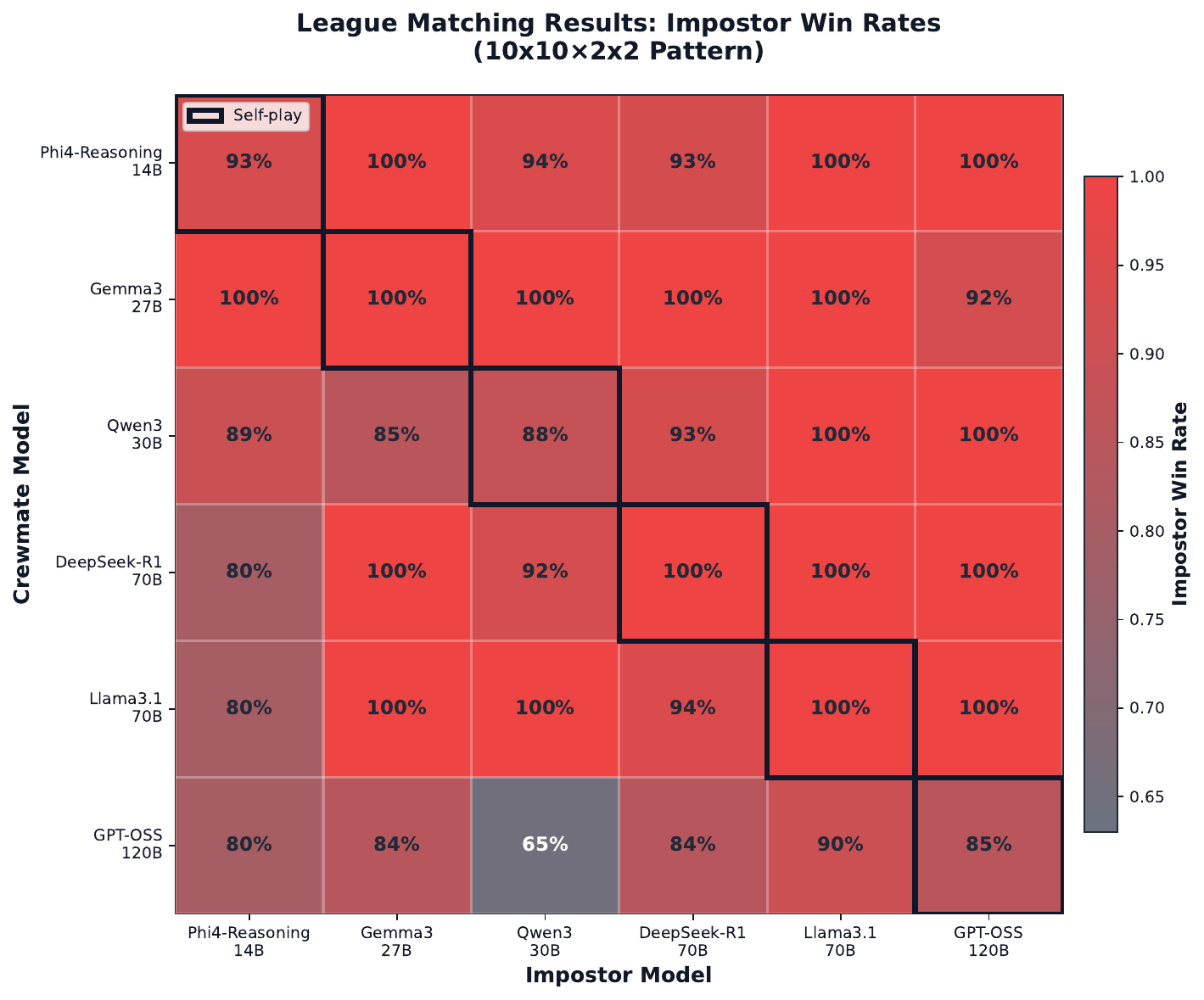}
    \end{minipage}
    \caption{\textbf{Head-to-Head Win Rate Heatmaps.} \textbf{Left:} Overall win rate---each cell shows the win rate of the row model against the column model on the $10{\times}10$\_2x2 pattern. Darker colors indicate higher win rates. The diagonal represents self-play and is set to 0.5. \textbf{Right:} Impostor win rate---complementary view showing impostor win rates for the same matchups. Colors range from gray (low) to red (high). The predominantly red coloring reflects the strong impostor advantage ($>80\%$ win rate).}
    \label{fig:league_heatmap}
\end{figure}

Figure~\ref{fig:league_heatmap} reveals two notable patterns. First, while larger models generally perform better, smaller models such as Qwen3-30B and Gemma3-27B occasionally outperform larger counterparts in specific matchups, suggesting that architecture and training objectives can partially compensate for scale. Second, GPT-OSS-120B exhibits the most consistent dominance across matchups, aligning with its top league ranking.

\section{Winning Score Calculation}
\label{app:winning_score_calculation}
We summarize the evolving ``who is ahead'' game situation with a single scalar that behaves like a crewmate win-likelihood over time. This quantity is not a learned predictor; it is a transparent heuristic designed to be comparable across models and episodes while remaining easy to interpret. We build it from three signals that align with the main mechanisms that decide outcomes in this environment. First, progress toward the crewmate objective should increase the crewmate advantage, so higher task completion increases the score. Second, social-deduction games are strongly influenced by numbers, so having more alive crewmates than impostors increases the score. Third, if crewmates systematically assign lower trust to impostors, this indicates that impostors are being identified, which should also increase the score. We then map the combined score through a smooth saturating nonlinearity so that weak evidence yields values near neutral, strong evidence saturates near certain win or loss, and the resulting curve can be read as a probability-like ``win-lean'' signal. Finally, to prevent late-time averages from being biased toward episodes that simply last longer, we also report a terminal-filled variant in which each episode contributes its eventual outcome after it ends, ensuring the aggregate curve reflects outcome tendencies rather than survival duration.

\paragraph{Winning rate in the transition (win-lean) plot.}
At each trust-snapshot time step $t$, we plot a heuristic estimate of the probability that the \emph{crewmate side} is leaning toward winning:
\begin{equation}
    p_{\text{crew}}(t)=\frac{1}{2}\Bigl(1+\tanh\bigl(\alpha\, s(t)\bigr)\Bigr)\in[0,1],
\end{equation}
where
\begin{equation}
    s(t)=w_{\text{task}}\bigl(\mathrm{TC}(t)-0.5\bigr)
    +w_{\text{alive}}\,\Delta_{\text{alive}}(t)
    +w_{\text{detect}}\bigl(\mathrm{DET}(t)-0.5\bigr).
\end{equation}
Here $\mathrm{TC}(t)\in[0,1]$ is the (normalized) task-completion rate, the alive-advantage term is
\begin{equation}
    \Delta_{\text{alive}}(t)=\frac{C(t)-I(t)}{C(0)},
\end{equation}
with $C(t)$ and $I(t)$ the numbers of alive crewmates and impostors, and the detection signal is
\begin{equation}
    \mathrm{DET}(t)=1-\overline{\mathrm{Trust}}_{\text{crew}\rightarrow\text{imp}}(t),
\end{equation}
i.e., one minus the mean trust score assigned by alive crewmates to alive impostors.

\paragraph{Terminal-filled variant (to reduce survivorship bias).}
After an episode ends at time $\tau$, we fill the remaining trajectory with the terminal outcome:
\begin{equation}
    \tilde p_{\text{crew}}(t)=
    \begin{cases}
        p_{\text{crew}}(t), & t\le \tau,                         \\
        1,                  & t>\tau \ \text{and crewmates win}, \\
        0,                  & t>\tau \ \text{and impostors win}.
    \end{cases}
\end{equation}
The plotted curve is the mean of $p_{\text{crew}}(t)$ (or $\tilde p_{\text{crew}}(t)$) across episodes for each model.

\begin{figure}[t]
    \centering
    \includegraphics[width=0.94\linewidth]{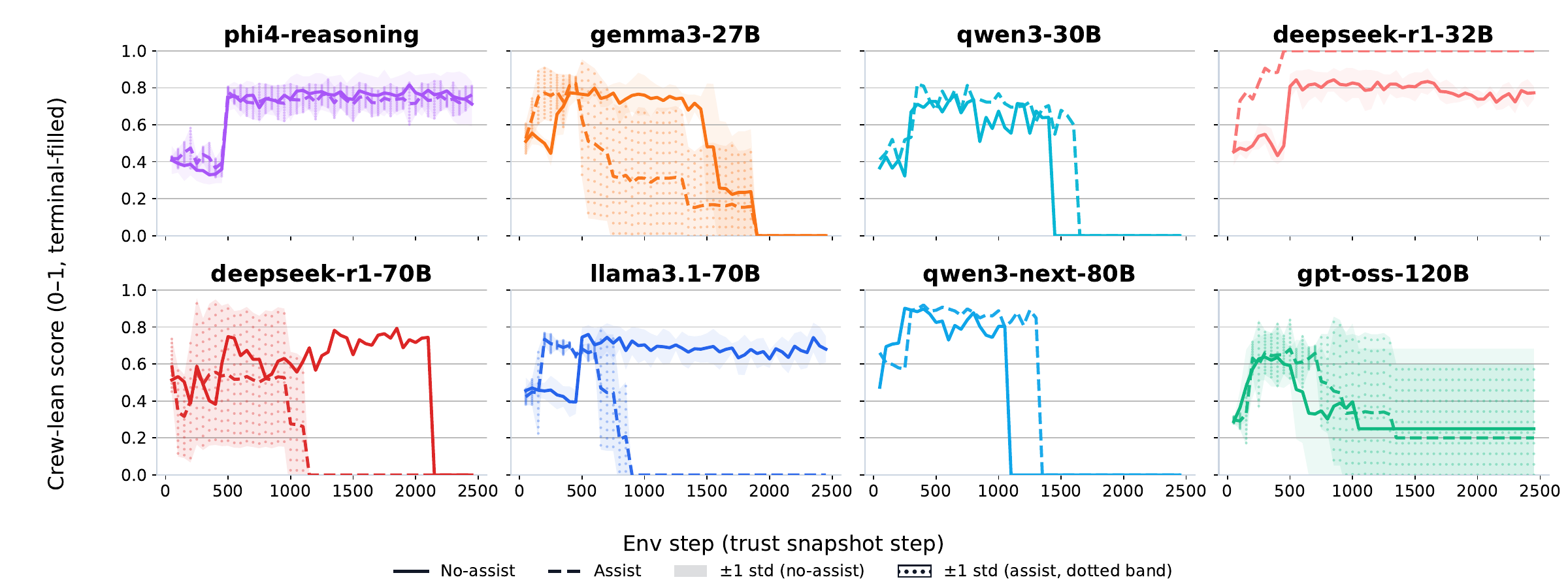}
    \caption{\textbf{Impostors gained advantage by the navigation assistant.} Comparison of winning score transition on \name.}
    \label{fig:winning_score_transition}
\end{figure}

\end{document}